\documentclass[opre,nonblindrev,copyedit]{informs3} 

\usepackage{pdfpages}
\usepackage{makecell}
\usepackage{ulem}
\DoubleSpacedXI
\OneAndAHalfSpacedXI
\usepackage{amsmath}
\usepackage{bm}
\usepackage{array}
\newcolumntype{P}[1]{>{\centering\arraybackslash}p{#1}}
\usepackage[framemethod=TikZ]{mdframed}
\usepackage{multicol}
\usepackage{endnotes}
\newcommand{\af}[1]{{\textcolor{red}{[AliF: #1]}}}
\newcommand{\mr}[1]{\textbf{\textcolor{purple}{[Meysam: #1]}}}

\newcommand{\aes}[1]{{\textcolor{blue}{[AliE: #1]}}}

\usepackage{tabularx}
\newcolumntype{Y}{>{\centering\arraybackslash}X}
\usepackage{longtable}

\usepackage{amsmath}
\usepackage{mathtools}
\usepackage{arydshln}
\usepackage{natbib}
 \bibpunct[, ]{(}{)}{,}{a}{}{,}%
 \def\bibfont{\small}%
\usepackage{tikz}
\usepackage{tikz-3dplot}
\usepackage{pgfplots}
\usetikzlibrary{shapes.geometric, arrows}
\usepackage{mathrsfs}
\usepackage{hyperref}
\TheoremsNumberedThrough     
\ECRepeatTheorems
\usepackage{algpascal}
\usepackage{algorithmicx}
\usepackage{algorithm}
\usepackage{algpseudocode}
\usepackage{anyfontsize}
\pgfplotsset{compat=1.16} 
\usetikzlibrary{pgfplots.statistics, pgfplots.colorbrewer} 
\usepackage{appendix}
\usepackage{enumitem}

\EquationsNumberedThrough    
\usepackage{pgf-pie}
\usepackage{xcolor,colortbl}
\definecolor{Tan}{rgb}{0.57, 0.51, 0.32}
\definecolor{amaranth}{rgb}{0.9, 0.17, 0.31}
\definecolor{asparagus}{rgb}{0.53, 0.66, 0.42}
\definecolor{babypink}{rgb}{0.96, 0.76, 0.76}
\definecolor{blush}{rgb}{0.87, 0.36, 0.51}
\definecolor{bazaar}{rgb}{0.6, 0.47, 0.48}
\definecolor{antiquefuchsia}{rgb}{0.57, 0.36, 0.51}
\definecolor{atomictangerine}{rgb}{1.0, 0.6, 0.4}
\definecolor{beaver}{rgb}{0.62, 0.51, 0.44}
\definecolor{brownt}{rgb}{0.59, 0.29, 0.0}
\definecolor{cadet}{rgb}{0.33, 0.41, 0.47}
\definecolor{anti-flashwhite}{rgb}{0.95, 0.95, 0.96}
\definecolor{ashgrey}{rgb}{0.7, 0.75, 0.71}
\definecolor{ballblue}{rgb}{0.13, 0.67, 0.8}
\definecolor{almond}{rgb}{0.94, 0.87, 0.8}
\definecolor{burntorange}{rgb}{0.8, 0.33, 0.0}
\usepackage{caption}
\usepackage{subcaption}
\usepackage{graphicx}
\usepackage{collcell}
\usepackage{diagbox}
\usepackage{rotating}

\usepackage{tabularx}
\newcolumntype{Y}{>{\centering\arraybackslash}X}

\usepackage{collcell}

\newcommand {\UU}  { {\bm{U}} }

\newcommand {\zz}  { {\bm z} }

\newcommand {\xx}  { {\bm x} }
\newcommand {\yy}  { {\bm y} }

\newcommand {\vv}  { {\bm v} }
\newcommand {\ww}  { {\bm w} }
\newcommand {\ff}  { {\bm f} }

\newcommand {\cc}  { {\bm c} }

\newcommand{\eq}{\mathrel{\mathop:}=}

\newcommand*\bigO[1]{\mathcal O\left( #1\right)}

\usepackage{mdframed}

\def\theARTICLETOPLEFT{}
\def\theARTICLETOPRIGHT{}

\begin{document}
\RUNAUTHOR{Fattahi, Eshragh, Aslani, and Rabiee}
\RUNTITLE{Ranking Vectors Clustering: Theory and Applications \vspace{-.25cm}}
\TITLE{\vspace{-3.0cm} \\ Ranking Vectors Clustering: Theory and Applications}

\ARTICLEAUTHORS{
\AUTHOR{Ali Fattahi}
\AFF{Carey Business School, Johns Hopkins University, Baltimore, MD USA, \EMAIL{ali.fattahi@jhu.edu} }
\AUTHOR{Ali Eshragh}
\AFF{Carey Business School, Johns Hopkins University, Washington, DC USA, \\ International Computer Science Institute, University of California at Berkeley, CA USA, \EMAIL{Ali.Eshragh@jhu.edu} }
\AUTHOR{Babak Aslani}
\AFF{Department of Systems Engineering and Operations Research, George Mason University, USA,
\EMAIL{baslani@gmu.edu} }
\AUTHOR{Meysam Rabiee}
\AFF{Business School, University of Colorado Denver, Denver, CO 80202, USA,
\EMAIL{meysam.rabiee@ucdenver.edu} }
}

\ABSTRACT{
We study the problem of clustering ranking vectors, where each vector represents preferences as an ordered list of distinct integers. Specifically, we focus on the $k$-centroids ranking vectors clustering problem (\texttt{KRC}), which aims to partition a set of ranking vectors into $k$ clusters and identify the centroid of each cluster. Unlike classical $k$-means clustering (\texttt{KMC}), \texttt{KRC} constrains both the observations and centroids to be ranking vectors. We establish the NP-hardness of \texttt{KRC} and characterize its feasible set. For the single-cluster case, we derive a closed-form analytical solution for the optimal centroid, which can be computed in linear time. To address the computational challenges of \texttt{KRC}, we develop an efficient approximation algorithm, \texttt{KRCA}, which iteratively refines initial solutions from \texttt{KMC}, referred to as the baseline solution. Additionally, we introduce a branch-and-bound (\texttt{BnB}) algorithm for efficient cluster reconstruction within \texttt{KRCA}, leveraging a decision tree framework to reduce computational time while incorporating a controlling parameter to balance solution quality and efficiency. We establish theoretical error bounds for \texttt{KRCA} and \texttt{BnB}. Through extensive numerical experiments on synthetic and real-world datasets, we demonstrate that \texttt{KRCA} consistently outperforms baseline solutions, delivering significant improvements in solution quality with fast computational times. This work highlights the practical significance of \texttt{KRC} for personalization and large-scale decision making, offering methodological advancements and insights that can be built upon in future studies.

\vspace{.2cm}

}

\maketitle

\section{Introduction}
\label{sec:introduction}

We study the problem of clustering ranking vectors for large-scale problems by developing theoretical frameworks and applying them to extensive synthetic and real datasets. A \textit{ranking vector} is an $m$-dimensional vector that represents a permutation of the set $\{1, \dots, m\}$. Each entry in the vector is unique, corresponding to exactly one element from the set. In a ranking vector, the value of each entry indicates the position or rank of the corresponding option in an ordered list \citep{Aledo2018, hou2019}. Note that there are no ties, as each position is assigned to exactly one element. For instance, an individual's order of preference for movie genres can be represented as a ranking vector, wherein each entry indicates the rank of the corresponding genre. Similarly, in a hiring committee, each member's preference ranking over applicants captures their individual evaluation of candidates. In a similar vein, students may rank their subject preferences based on interest or perceived difficulty, while in healthcare, patients may rank treatment options according to factors such as effectiveness, side effects, and cost. These examples illustrate how ranking vectors naturally arise in diverse decision making contexts.

The \textit{$k$-centroids ranking vectors clustering} problem, denoted by \texttt{KRC}, involves partitioning a given set of $n$ ranking vectors into $k$ clusters and determining the centroid of each cluster by minimizing the sum of the squared Euclidean distances between the $n$ ranking vectors and their assigned centroids. While \texttt{KRC} may seem similar to the classical $k$-means clustering (\texttt{KMC}) problem \citep{lloyd1982least, Arthur2006}, where both data points and centroids can take any real values, \texttt{KRC} imposes a stricter constraint: both the data points and the centroids must be ranking vectors. This structured discretization of the decision variables introduces new challenges to the problem.

The \texttt{KRC} problem enables the personalization of offers{, promotions,} and recommendations by clustering ranking vectors, making it applicable across various domains such as enhancing user experiences on digital platforms. Beyond personalization, it also plays a crucial role in decision making contexts where diverse perspectives must be aggregated, such as {in group decision making}. Given its ability to structure ranking data efficiently, \texttt{KRC} is particularly valuable for large-scale applications, where $n$ can scale to millions, requiring scalable and efficient computational methods.

In Section \ref{subsec:MotivatingExamples}, we propose several applications for \texttt{KRC} to emphasize its broad applicability and discuss how our proposed methodology can potentially lead to significant improvements in these applications.
We proceed by presenting our motivating examples in Section \ref{subsec:MotivatingExamples} and outlining our main results and insights in Section \ref{subsec:summaryofresults}.

\subsection{Motivating Examples} 
\label{subsec:MotivatingExamples}

We propose several applications for \texttt{KRC} across diverse domains, ranging from online review platforms—illustrated by examples {including streaming platforms such as Netflix, Amazon Prime, and HBO Max, hospitality review systems, } 
and dining reservation platforms like OpenTable and Resy—to large-scale group decision making. To the best of our knowledge, the application of ranking vector clustering in these contexts is novel and has not been explored in the existing literature, as detailed in Section \ref{sec:LiteratureReview&Background}.

\subsubsection{Online Review Platforms.} 
\label{subsubsec:online_review_platforms}
On streaming platforms such as Netflix, Amazon Prime, and HBO Max, viewers often rely on the average of {individual ratings to decide what to watch.} However, differences in genre preferences can result in significant variation in ratings, leading to high volatility. To address this, we propose a personalized approach to presenting average ratings that incorporates viewers’ genre preference rankings. While prior work has explored personalization based on viewer characteristics (Section \ref{sec:LiteratureReview&Background}), to the best of our knowledge, we are the first to personalize ratings using genre preference rankings. This approach is motivated by our hypothesis that differences in genre preferences drive variation in movie ratings.

For instance, consider two types of viewers: A and B. Viewer A prefers comedy over thriller, while viewer B favors thriller over comedy. Suppose type A viewers consistently assign higher ratings to comedies, and type B viewers do the same for thrillers, reflecting the greater enjoyment they derive from their preferred genres. We explore such systematic rating behaviors using the MovieLens dataset (\href{https://grouplens.org/datasets/movielens/}{grouplens.org/datasets/movielens}) in Section \ref{sec:numerical_experiments}. If a type A viewer relies on average ratings to choose between a comedy and a thriller, they might end up selecting a less desirable option. This misalignment is particularly likely when type A viewers are a minority --- that is, when the number of type B viewers significantly outweighs that of type A. 

Research suggests a positive relationship between customer satisfaction and profitability \citep{pooser2018effects, williams2011customer}. Building on this insight, personalization based on genre preference rankings has the potential to significantly enhance viewer utility, particularly for minority groups. This personalization can be implemented using \texttt{KRC} to identify clusters of viewers with similar genre preference rankings. Based on these clusters, we propose that the platform personalize the presentation of average ratings. Specifically, a viewer would see the average rating assigned by others within the same cluster, reflecting shared genre preferences. 

Other practical applications of online review platforms include the hospitality industry and dining reservation platforms. In these contexts, a set of \textit{criteria} plays a critical role in shaping customer satisfaction. For instance, in online accommodation booking platforms like Airbnb or TripAdvisor, the criteria might include cost, location, amenities, cleanliness, and proximity to public transportation. A budget-conscious traveler may prioritize cost over location, whereas a wealthier traveler might value location more highly. Similarly, in dining reservation platforms such as OpenTable and Resy, criteria may include cost, menu variety, service quality, location, and reservation availability.

We propose a personalized approach based on criteria preference rankings for these contexts. First, the platform determines each customer's criteria preference ranking, either directly (e.g., by surveying subscribers) or indirectly (e.g., through contextual data analytics) \citep{panniello2016context,shen2023learning}. The \texttt{KRC} problem is then applied to form clusters of customers with similar preference rankings. {Finally, the platform could personalize the presentation of average ratings, as well as tailor offers, promotions, and recommendations for each cluster.}

As exemplified in the above applications, \texttt{KRC} has significant applications in customer segmentation for personalized and customized products and services, as well as in data summarization and machine learning. It enables targeted marketing, promotions, and tailored recommendations across these domains, leading to better decisions.

\subsubsection{Large-scale Group Decision Making.} 
\label{subsubsec:large-scale_group_dm}
Group decision making involves a process in which multiple decision makers participate, and a final solution is required that reflects the collective knowledge and preferences of the group \citep{altuzarra2010consensus}. We are particularly focused on large-scale applications, where millions of decision makers could be involved. For example, in urban planning, responding to criticisms regarding the lack of inclusivity and limited public input, involving citizens in the decision making process has become increasingly popular \citep{ertio2015participatory}.

The application of large-scale group decision making is expanding rapidly due to the growth of social networks, internet-based systems, and the increasing recognition of new paradigms such as e-democracy \citep{Kim2008} and international trade law unification efforts, leading to decision making problems involving dozens to thousands of decision makers \citep{Tang2020}. Recently, large-scale group decision making has been applied in a variety of fields, including water management \citep{Srdjevic2007}, emergency decision making \citep{Xu2015}, education management \citep{Rodriguez2018}, social credit systems \citep{Ma2019}, financial inclusion \citep{chao2020}, hospital ranking \citep{liao2020}, green enterprise ranking problems, and the excavation scheme selection problem for shallow buried tunnels \citep{liao2020DNMA}.

Consider a scenario where the citizens of a metropolitan area have provided their preference rankings for four areas of development: public transportation, affordable housing, recreational facilities, and energy efficiency. These four areas lead to $24$ unique preference rankings. Citizens often form diverse clusters, each with significant differences that influence how they evaluate and compare these options. For instance, one cluster may prefer improving public transportation over recreational facilities, while another cluster may favor the opposite. Ignoring these clusters could result in a decision that is not representative of any specific group, nor the overall population.

By applying \texttt{KRC}, we identify these clusters and determine a centroid for each one. A centroid can be viewed as the proposal that best represents the preferences of its corresponding cluster. Essentially, a centroid summarizes the collective knowledge and preferences of the group it represents. With the clusters and their centroids identified, a final decision making committee (such as a small group of experts) can make an informed decision, referred to as the consensus reaching process \citep{liang2022consensus}. Identifying clusters also offers additional benefits; for example, mapping them onto geospatial areas can help customize city development plans, providing a more tailored approach rather than a one-size-fits-all solution.


In summary, we illustrated the broad applicability of \texttt{KRC} through examples in online platforms and large-scale group decision making. In online platforms, where customers' criteria preference rankings influence their decisions, identifying clusters with homogeneous rankings enables personalized {presentations, promotions, and recommendations, improving user satisfaction and profitability.} In large-scale group decision making, clustering ranking vectors helps derive consensus solutions that reflect the preferences of all groups. The \texttt{KRC} framework unifies these applications, highlighting the need for effective solution approaches.

\subsection{Summary of Main Results}
\label{subsec:summaryofresults}

Our contributions are outlined below:

\begin{enumerate}[label=\ \ \ \ \ (\roman*)\ \ , align=left, leftmargin=0pt, labelsep=0em, itemsep=0ex, listparindent=0pt]
    \item We present the first comprehensive theoretical study of the ranking vectors clustering problem. We formulate \texttt{KRC} mathematically, prove that it is \texttt{NP}-hard, and examine its similarities and differences with the $k$-means clustering problem. In addition, we show that if a clustering solution for \texttt{KMC} is $\varepsilon$-optimal, it will yield a $(1+2\varepsilon)$-optimal solution for \texttt{KRC}. This implies that a solution to \texttt{KMC} can be far from the \texttt{KRC} solution, as confirmed in our numerical experiments, emphasizing the importance of the theoretical development and applications of \texttt{KRC} presented in this paper.

    \item We provide a characterization of the feasible set for \texttt{KRC}, derive an analytical closed-form solution to determine the optimal centroid for each cluster, and show that it can be computed in linear time relative to the number of observations and options.

    \item We develop an approximation algorithm for \texttt{KRC}, called \texttt{KRCA}. This algorithm starts with an initial solution derived from \texttt{KMC} and iteratively refines it through a local search process. During each iteration, the \texttt{KRCA} algorithm updates the centroids and reassigns the observation ranking vectors to the nearest centroids. Each iteration leverages the unique structure of ranking vectors to enhance computational efficiency. {Notably, if the initial \texttt{KMC} solution is $\varepsilon$-optimal, \texttt{KRCA} produces a $(1+2\varepsilon)$-optimal solution.}

    \item We develop a branch-and-bound algorithm, denoted by \texttt{BnB}, for reconstructing the clusters at each iteration of the \texttt{KRCA} algorithm. The \texttt{BnB} algorithm employs a decision tree approach to partition the ranking vectors into subsets, represented as nodes in the tree. Within each node, the \texttt{BnB} algorithm updates lower and upper bounds for each centroid, enabling the comparison and elimination of non-promising centroids. Nodes are pruned if they contain no ranking vectors or only one centroid. We show that the \texttt{BnB} algorithm offers several advantages over an {exhaustive search (\texttt{ES}) method,} significantly reducing computation time for reconstructing clusters in instances with small $m$ and large $n$. Additionally, it incorporates an error factor $\epsilon$, allowing for sub-optimal cluster reconstruction with a worst-case error of $n(k-1)\epsilon$, which balances solution quality and computational efficiency. Furthermore, it leverages the structure of the input ranking vectors, reducing computation time when the input is inherently clustered. We theoretically investigate the \texttt{BnB} algorithm's convergence and establish its worst-case error. Additionally, we provide a bound on the depth of the decision tree as a function of the clusteredness of the observation ranking vectors.

    \item We conduct a comprehensive numerical analysis to assess the performance of the \texttt{KRCA} algorithm, using both simulated synthetic data and real datasets. Our simulations include instances with uniformly generated ranking vectors as well as those where the ranking vectors are inherently clustered. Additionally, we utilize a dataset from MovieLens, develop a procedure for creating benchmark instances from this dataset, and apply the \texttt{KRCA} algorithm to these benchmark instances. Our results demonstrate that the \texttt{KRCA} algorithm is both computationally efficient and provides substantial improvements compared to benchmark solutions in both simulated and real instances. Furthermore, our numerical results reveal that there are situations in which a solution to the \texttt{KMC} problem can be far from the \texttt{KRC} problem solution.
\end{enumerate}

The remainder of this paper is organized as follows. Section \ref{sec:LiteratureReview&Background} {reviews the} related literature. Section \ref{sec:model_formulation_properties} defines and formulates the \texttt{KRC} problem and establishes its \texttt{NP}-hardness. Section \ref{sec:single-cluster_error_bounds} introduces a closed-form solution for the single-cluster case and further develops the theoretical properties of the \texttt{KRC} problem. Section \ref{sec:krca} builds on these theoretical foundations to propose {the \texttt{KRCA} and \texttt{BnB} algorithms}, along with their analytical properties. Section \ref{sec:numerical_experiments} presents extensive numerical experiments on synthetic and real-world datasets to evaluate the computational performance and solution quality of \texttt{KRCA}. Finally, Section \ref{sec:conclusion_future_work} summarizes the key findings and outlines potential directions for future research. All proofs are presented in Online Appendix \ref{sec:appendix}.


\textbf{Notation.} Throughout this paper, vectors are represented by bold lowercase letters (e.g., $\vv$), while scalars are denoted by regular lowercase letters (e.g., $c$). All vectors are row vectors, {and $\langle \vv, \ww \rangle$ represents the inner product of $\vv$ and $\ww$.} For a real vector $\vv$, $\| \vv \|$ denotes the $\ell_2$ norm. Using $\mathtt{MATLAB}$ notation, $\vv(i:j)$ refers to a subvector of $\vv$ containing the $i^{\text{th}}$ to $j^{\text{th}}$ (inclusive) coordinates.
 

\section{Literature Review}
\label{sec:LiteratureReview&Background}

Clustering is a fundamental problem in data analysis with diverse theoretical and practical applications. Among its many variants, $k$-means clustering has gained considerable attention due to its simplicity and effectiveness in partitioning data~\citep{macqueen1967some,ikotun2023k}. A discrete variation, known as discrete $k$-means clustering (\texttt{DKMC}), introduces additional constraints by restricting both data points and centroids to a finite, predefined set of values. In this setting, inputs and cluster representatives are categorical or ordinal, such as labels or rankings, rather than continuous numerical values. Like \texttt{KMC}, \texttt{DKMC} also allows for ties, where multiple elements may share the same value~\citep{huang1998extensions}.

In addition to classical $k$-means, several other variants have been developed to handle diverse data types and clustering objectives. One such method is $k$-medoids clustering~\citep{kaufman1990finding,he2024crowd}, which replaces centroids with actual data points (medoids). Another variant, known as $k$-medians clustering~\citep{jain1988algorithms}, uses the median and the rectangular distance metric, making it more suitable for data with non-Euclidean geometry or ordinal structure. For purely categorical data, algorithms such as $k$-modes~\citep{huang1998extensions} and $k$-histograms~\citep{he2005k} replace the mean with a mode or frequency-based representation, respectively.

The \texttt{KRC} problem extends \texttt{KMC} and \texttt{DKMC} to scenarios where both observation vectors and centroids are ranking vectors. These ranking vectors are permutations of integers where no ties are allowed, meaning every element must hold a unique position. This constraint adds significant complexity to the problem, requiring new theoretical insights and algorithmic innovations to address its unique challenges and applications effectively.

\subsection{Theory and Algorithms} 
\label{subsec:lit_theory&algo}

The \texttt{KRC} problem is closely related to the classical \texttt{KMC} problem and its discrete version \texttt{DKMC}. The \texttt{KMC} problem aims to partition a set of $n$ observation points into $k$ clusters by minimizing the sum of squared Euclidean distances between each point and the centroid of its assigned cluster \citep{ikotun2023k}. Each centroid in \texttt{KMC} is computed as the mean of the observation points within the corresponding cluster. By contrast, \texttt{DKMC} introduces the constraint that centroids must be chosen from the input data points \citep{aggarwal2009adaptive}. Distinguishing itself from both \texttt{KMC} and \texttt{DKMC}, \texttt{KRC} requires that both the observations and centroids are ranking vectors. Both \texttt{KMC} and \texttt{DKMC} are known to be \texttt{NP}-hard \citep{dasgupta2008hardness,matouvsek2000approximate}. In this paper, we establish the \texttt{NP}-hardness of \texttt{KRC} via a reduction from the hypercube clustering problem, where binary vectors are clustered, and each cluster's centroid is restricted to be a binary vector \citep{baldi2012boolean}.

Given the challenges of solving \texttt{KMC} optimally, many existing studies focus on developing approximation algorithms. \citet{macqueen1967some} introduced the term ``$k$-means'' and proposed an iterative method that initializes $k$ random points as clusters. At each iteration, points are added to the cluster with the nearest mean (centroid), and the mean is subsequently updated. \citet{bandyapadhyay2015approximate} developed a bi-criteria polynomial-time approximation scheme (PTAS) for \texttt{KMC}, which identifies $(1+\varepsilon)k$ centers with an objective value at most $(1+\varepsilon)$ times the optimal solution. However, whether a true PTAS exists for \texttt{KMC} in the plane remains an open question. \citet{kanungo2002local} presented a polynomial-time approximation algorithm with a worst-case relative error of $(9+\varepsilon)$, which has since been improved to a PTAS for fixed dimensions by \citet{friggstad2019exact, friggstad2019local} and \citet{cohen2019local}. For arbitrary dimensions, \citet{ahmadian2019better} proposed a $6.357$-approximation. Our approximation algorithm, \texttt{KRCA}, builds upon these advancements by starting from a \texttt{KMC}-based initial solution. We prove that if the \texttt{KMC} solution is $\varepsilon$-optimal, then \texttt{KRCA} achieves $(1+2\varepsilon)$-optimality for the \texttt{KRC} problem.

Various adaptations of \texttt{DKMC} have been proposed to handle categorical, binary, or ordinal data. Techniques for solving \texttt{DKMC} include enumeration, integer programming, linear programming, and dynamic programming \citep{hansen1997cluster, jensen1969dynamic, vinod1969integer, rao1971cluster}. Lagrangian relaxation methods have also been used for datasets with fewer than 200 data points \citep{mulvey1979cluster}. However, these methods are unsuitable for \texttt{KRC} because it does not restrict centroids to be chosen from observations and is instead designed for large-scale datasets. 

Lloyd's algorithm is a widely used local search method for \texttt{KMC} due to its simplicity and effectiveness \citep{lloyd1982least}. The algorithm alternates between assigning each observation point to the nearest centroid and updating centroids by calculating the cluster means. The process repeats until a stopping criterion is satisfied. Lloyd's approach is closely related to the expectation-maximization method \citep{dempster1977maximum}. Several improved versions of Lloyd's algorithm, such as $k$-means$++$ \citep{Arthur2006}, have been proposed to enhance the initialization process. Our \texttt{KRCA} algorithm shares a similar framework but adapts each step to exploit the structure of ranking vectors. We introduce a polynomial-time procedure for calculating the optimal centroid in initialization and update steps and propose a novel \texttt{BnB} algorithm for the assignment step. {We present} theoretical analyses and derive error bounds for both the \texttt{KRCA} and \texttt{BnB} algorithms.

{Similar to \texttt{BnB},} \citet{kanungo2002efficient} utilized a decision tree for assigning observations to centroids in \texttt{KMC}. However, our \texttt{BnB} algorithm introduces a new branching scheme, bounding procedures within tree nodes, and theoretical analyses of error and tree depth. These innovations leverage the unique properties of ranking vectors, making our approach distinct and well-suited to \texttt{KRC}.

\subsection{Applications}
\label{subsec:lit_application}

As discussed in Section \ref{subsec:MotivatingExamples}, \texttt{KRC} has broad applications in online review platforms and large-scale group decision making. This section reviews the literature on these topics and highlights how \texttt{KRC} contributes to these domains.

\subsubsection{Online Review Platforms.} 
\label{subsubsec:online_review_platforms}
The growth of e-commerce has made customer feedback, especially in the form of online reviews, a crucial resource for analyzing market trends and tailoring individual customer experiences \citep{chen2008online}. Online product reviews are the second most trusted source of information for consumers, following referrals from friends and family \citep{nielsen2012global}. According to the Pew Research Center’s 2016 Online Shopping and E-Commerce survey \citep{pew2016onlineshopping}, $82\%$ of Americans consult reviews before making a first-time purchase. This highlights the dual importance of reviews for customers making informed decisions and businesses aiming to understand consumer preferences \citep{cui2012effect}.

A significant body of research focuses on understanding customer preferences through online reviews, particularly by identifying key product features. Early works, such as \citet{godes2004using} and \citet{liu2006word}, utilized human coding to analyze textual data. Recent advancements leverage natural language processing techniques to extract insights methodically \citep{netzer2012mine, tirunillai2014mining}. Studies also incorporate demographic and geographic factors to personalize offers and recommendations \citep{raoofpanah2023review}. However, to the best of our knowledge, no existing research employs feature rankings {as a foundation for  personalization}. The \texttt{KRC} problem addresses this gap by {clustering customers} based on ranking vectors, offering a novel approach for e-commerce applications.

\subsubsection{Large-scale Group Decision Making.} 
\label{subsubsec:large-scale_group_dm}
Large-scale group decision making often involves solving vector-clustering problems, which can be classified based on their distance measures, input vector types, and centroid representations. Common distance measures include squared Euclidean distance \citep{Tang2019b}, Kendall-tau distance \citep{Farnoud2012}, and Spearman’s footrule \citep{Chen2011}. Input vectors and centroids studied in the literature include ranking vectors \citep{Aledo2018,hou2019}, partial ranking vectors \citep{amodio2016,Dinu2006}, and score vectors \citep{de2002}. {We focus exclusively on clustering ranking vectors using squared Euclidean distance as the distance measure.} 

In single-cluster scenarios, this problem is referred to as \textit{rank aggregation} in the group decision making literature \citet{xiao2017}. Borda-based heuristics are widely used for determining aggregate rankings \citep{balinski2014judge, zhou2024heuristic}. These methods typically involve converting rankings to scores and aggregating the scores. Recent advances, such as \citet{zhou2024heuristic}, propose hybrid evolutionary algorithms that effectively handle both complete and partial rankings, using Borda-based heuristics as a starting point. \citet{Brandl2019} offers a comprehensive study of the properties of these approaches. For further developments in single-cluster rank aggregation, see \citet{Yoo2020,Aledo2018,Caragiannis2019,Dopazo2017,Chatterjee2018,Chin2004}, and \citet{Raisali2013}.

Our contribution to this domain includes developing a polynomial-time procedure for solving \texttt{KRC} in the single-cluster case, which determines an aggregate ranking vector that minimizes the squared Euclidean distance. While existing rank aggregation methods, such as those by \citet{Schalekamp2009,Farnoud2012,Dinu2006,Ding2018h,Negahban2017}, focus on the single-cluster case of \texttt{KRC}, they employ different objective functions or input types and are therefore not directly applicable to our setting. For instance, \citet{Schalekamp2009} minimizes the Kendall-tau distance, while \citet{Farnoud2012} introduces an approximation method based on Spearman's footrule distance. \citet{Dinu2006} explores rank-distance, a measure for partial rankings, and \citet{Ding2018h} develops a hierarchical heuristic for combining rankings with ambiguous preferences. \citet{Negahban2017} proposes Rank Centrality, an iterative algorithm that infers scores from pairwise comparisons. In contrast, our approach focuses on full rankings and minimizes the squared Euclidean distance, making it distinct from these prior methods.

For multiple-cluster cases, the literature is relatively sparse. Existing methods are primarily heuristic-based \citep{Zhou2020,Wu2020,Tang2019b}. The \texttt{KRC} problem contributes to this area by introducing an effective approximation algorithm for multi-cluster problems. 

{We provide a thorough theoretical characterization for the \texttt{KRC} problem, prove \texttt{NP}-hardness, develop the \texttt{KRCA} approximation algorithm, and establish error bounds.}


\section{\texttt{KRC}: Mathematical Formulation and Complexity Analysis}
\label{sec:model_formulation_properties}

In this section, we mathematically formulate the \texttt{KRC} problem and rigorously develop its analytical properties. Furthermore, we prove that \texttt{KRC} is \texttt{NP}-hard and highlight its distinctions from \texttt{KMC}. {To establish a clear foundation for our analysis, we begin by formally defining the \texttt{KMC} problem.}

\begin{definition}[$k$-Means Clustering \citep{lloyd1982least,Arthur2006}] 
    \label{def:kmc}
    Consider a set of $n$ observations $\mathcal{S} = \{\xx_1,\ldots, \xx_n \}$, where $\xx_i \in \mathbb{R}^m$ for $i=1,\ldots,n$. The \textit{$k$-means clustering} problem, denoted by \texttt{KMC}, is defined as finding a partition of $\mathcal{S}$ into $k$ mutually exclusive and collectively exhaustive subsets $\mathcal{S}_1, \dots, \mathcal{S}_{k}$ and centroids $\yy_1, \dots, \yy_{k} \in \mathbb{R}^m$ such that the objective function \begin{align} 
        v_{\mathtt{KMC}}(\mathcal{S}_1, \dots, \mathcal{S}_{k},\yy_1,\ldots,\yy_k) & \eq \sum_{\ell=1}^{k}\sum_{\xx\in\mathcal{S}_{\ell}} \| \xx - \yy_{\ell} \|^2 
        \label{eq:objFn_kmc} 
    \end{align} 
    is minimized, where $\|\cdot\|^2$ denotes the squared Euclidean distance. Let $v_{\mathtt{KMC}}^*$ denote the optimal value of the objective function in Equation \eqref{eq:objFn_kmc} for the \texttt{KMC} problem. 
\end{definition}

It is well established \citep{lloyd1982least,Arthur2006} that for any given clusters $\mathcal{S}_1, \dots, \mathcal{S}_{k}$, the objective function \eqref{eq:objFn_kmc} is minimized if the centroids are set to the mean of all observations in the clusters, expressed as: 
\begin{align}
    \label{eq:optimal_center_kmc} 
    \yy_\ell^* & = \frac{\sum_{\xx \in \mathcal{S}_\ell} \xx}{| \mathcal{S}_\ell |} \quad \mbox{for } \ell=1,\ldots,k, 
\end{align} 
where $ |\mathcal{S}_\ell| $ is the cardinality of the set $\mathcal{S}_\ell$ (assuming $ |\mathcal{S}_\ell| \neq 0$). Consequently, solving the \texttt{KMC} problem reduces to finding the optimal clusters. For simplicity, the centroids $\yy_1,\ldots,\yy_k$ are omitted from the notation of the objective function \eqref{eq:objFn_kmc}, and it is denoted as $v_{\mathtt{KMC}}(\mathcal{S}_1, \dots, \mathcal{S}_{k})$, implying that the centroids are optimally determined according to Equation \eqref{eq:optimal_center_kmc}.

Next, we introduce the \texttt{KRC} problem. Consider $m$ distinct options, such as the criteria or candidates discussed in Section \ref{subsec:MotivatingExamples}. Suppose these options are ranked by individuals with no ties, from the least preferred ($1^\text{st}$) to the most preferred ($m^\text{th}$). The output provided by each individual is referred to as a ranking vector, formally defined as follows.

\begin{definition}[Ranking Vector] A vector $\xx \in \mathbb{R}^m$ is called a \textit{ranking vector} if it is a permutation of the integers $1$ to $m$. Additionally, let $\mathcal{X}_m$ denote the set of all $m!$ possible {ranking vectors.} \end{definition}

{We provide an intuitive geometric representation for the space of all $m$-dimensional ranking vectors $\mathcal{X}_m$.} We demonstrate that ranking vectors lie at the intersection of a specific hypersphere and hyperplane. Let $\mathsf{BHS}(\zz, r^2) = \{\xx \in \mathbb{R}^m \mid \| \xx - \zz \|^2 = r^2 \}$ represent the boundary of a hypersphere centered at $\zz \in \mathbb{R}^m$ with radius $r \in \mathbb{R}^{+}$. Similarly, let $\mathsf{HP}(\yy, c) = \{\xx \in \mathbb{R}^m \mid \langle \yy, \xx \rangle = c \}$ denote a hyperplane with a normal vector $\yy \in \mathbb{R}^m$ and a constant $c \in \mathbb{R}$. Lemma \ref{lem:characterization_ranking_vectors} characterizes and provides a geometric perspective of $\mathcal{X}_m$.

\begin{lemma}[Characterization of Ranking Vectors] 
    \label{lem:characterization_ranking_vectors} 
    The set of all $m$-dimensional ranking vectors $\mathcal{X}_m$ satisfies the following inclusion equation:
    \begin{align*} 
        \mathcal{X}_{m} &\subseteq \mathsf{BHS}\left(\mathbf{0},\frac{1}{6}m(m+1)(2m+1)\right) \cap\ \mathsf{HP}\left(\mathbf{1},\frac{1}{2}m(m+1)\right), 
    \end{align*} 
    where $\mathbf{0}$ and $\mathbf{1}$ are vectors of size $m$ with all entries equal to zero and one, respectively.
\end{lemma}

\begin{figure}[t] 
    \centering 
    \includegraphics[width= 0.35\textwidth]{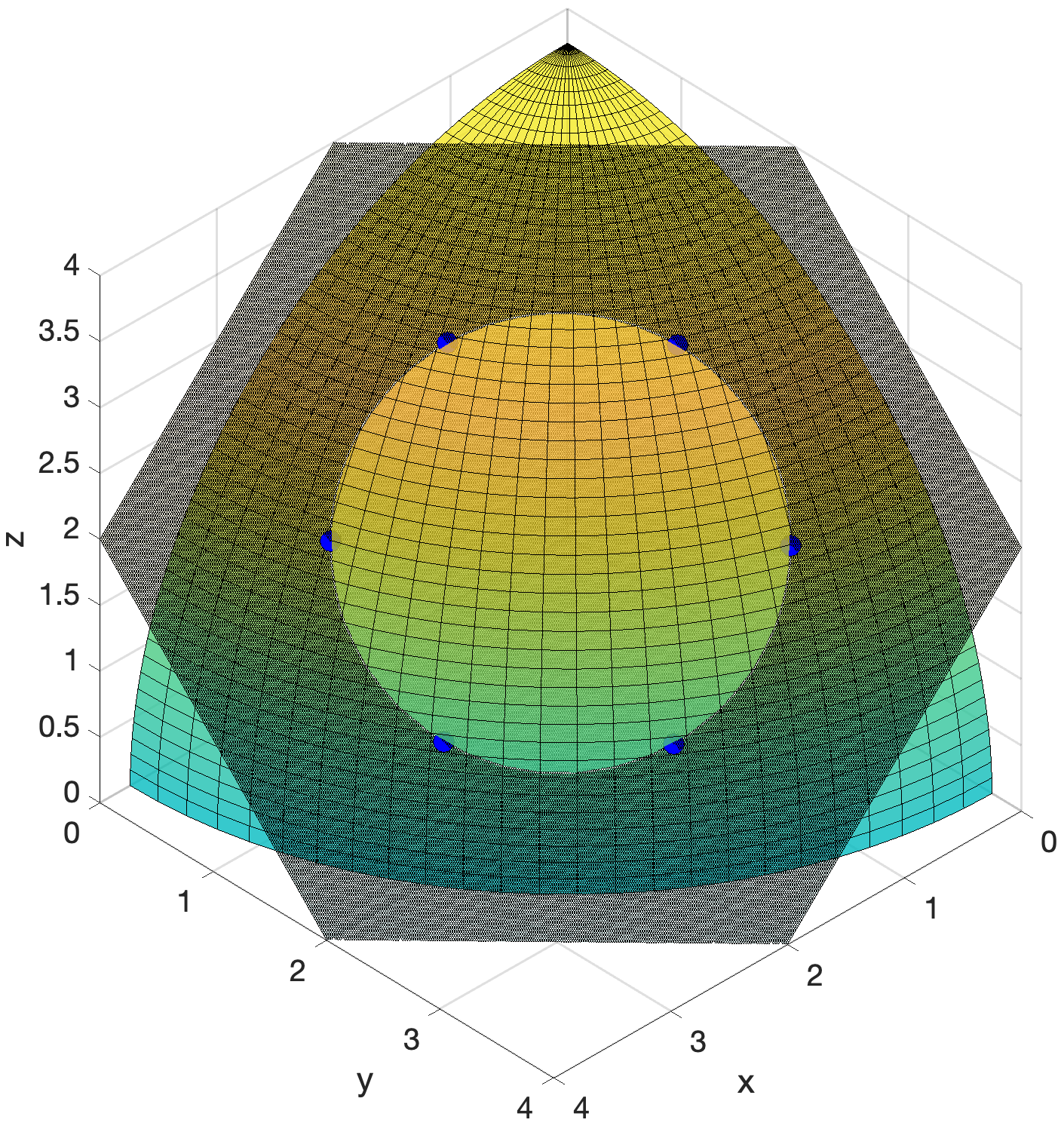} 
    \caption{The six round blue points demonstrate the set $\mathcal{X}_3$ as per the inclusion equation given in Lemma \ref{lem:characterization_ranking_vectors}.} 
    \label{fig:x_pr} 
\end{figure}

Lemma \ref{lem:characterization_ranking_vectors} establishes that ranking vectors reside at the intersection of a specific hypersphere and hyperplane. Figure \ref{fig:x_pr} illustrates this inclusion equation for the case $m=3$. The six round blue points correspond to the ranking vectors $\begin{bmatrix}1 & 2 & 3\end{bmatrix}$, $\begin{bmatrix}1 & 3 & 2\end{bmatrix}$, $\begin{bmatrix}2 & 1 & 3\end{bmatrix}$, $\begin{bmatrix}2 & 3 & 1\end{bmatrix}$, $\begin{bmatrix}3 & 1 & 2\end{bmatrix}$, and $\begin{bmatrix}3 & 2 & 1\end{bmatrix}$. The intersection of the hypersphere and the hyperplane forms a circle in $\mathbb{R}^3$, and the six points lie precisely on the circle's boundary. Notably, $\mathcal{X}_{m}$, and even the circle itself, occupy only a small portion of $\mathbb{R}^3$.

With this geometric intuition established, we now formally define the \texttt{KRC} problem. Consider $n$ individuals and $m$ options, with $\xx_i$ representing the ranking vector assigned by individual $i$ to the $m$ options, where $i = 1, \ldots, n$.
\begin{definition}[$k$-Ranking Clustering]
    \label{def:krc} 
    Let $\mathcal{S} = \{\xx_1,\ldots, \xx_n\}$, where $\xx_i \in \mathcal{X}_m$ for $i=1,\dots,n$, denote the set of all ranking vectors provided by $n$ individuals. The \textit{$k$-ranking clustering} problem, denoted as \texttt{KRC}, is defined as finding a partition of $\mathcal{S}$ into $k$ mutually exclusive and collectively exhaustive subsets $\mathcal{S}_1, \dots, \mathcal{S}_{k}$ and centroids $\yy_1, \dots, \yy_{k} \in \mathcal{X}_m$, such that 
    \begin{align} 
        v_{\mathtt{KRC}}(\mathcal{S}_1, \dots, \mathcal{S}_{k},\yy_1,\ldots,\yy_k) & \eq \sum_{\ell=1}^{k}\sum_{\xx\in\mathcal{S}_{\ell}} \| \xx - \yy_{\ell} \|^2 
        \label{eq:objFn_krc} 
    \end{align} 
    is minimized.\footnote{Alternative distance measures may be applicable in certain contexts. However, the squared Euclidean distance is used here as it appropriately penalizes large differences. This approach aligns with the normalized squared Euclidean distance measure employed by \citet{merigo2011} in the context of large-scale group decision making.} Let $v_{\mathtt{KRC}}^*$ denote the optimal value of the objective function in Equation \eqref{eq:objFn_krc} for the \texttt{KRC} problem. 
\end{definition}

We now examine the computational challenges of \texttt{KRC}. Intuitively, the complexity of \texttt{KRC} is influenced by the parameters $n$, $m$, and $k$. As discussed in Section \ref{subsec:MotivatingExamples}, for big-data applications—the primary focus of this paper—$n$ can reach the order of millions. In contrast, $m$ and $k$ may be either small or large, depending on the application. Consequently, developing a solution approach that is both scalable for large $n$ and adaptable to varying values of $m$ and $k$ is essential. While integer programming techniques may be effective for small instances of \texttt{KRC}, their applicability diminishes for real, large-scale instances, as outlined in the following remark.

\begin{remark}[Integer Program for \texttt{KRC}]\label{rem:integer_program} 
    The \texttt{KRC} problem can be formulated as an integer program by (i) enumerating a cost matrix, where rows represent the $n$ ranking vectors provided by individuals and columns represent the $m!$ possible centroids, and (ii) modeling the problem as selecting $k$ centroids and assigning the input vectors to these centroids. This formulation resembles the classical {\it p-median} problem \citep{daskin2015p} and is computationally feasible only for small values of $n$, $m$, and $k$. 
\end{remark}

As highlighted in Remark \ref{rem:integer_program}, exact methods such as integer programming are impractical for solving large-scale instances of \texttt{KRC}. While the geometric structure of ranking vectors, as characterized in Lemma \ref{lem:characterization_ranking_vectors}, might indicate potential simplifications, the computational complexity remains substantial. For example, leveraging the insights from Remark \ref{rem:integer_program}, the special case of \texttt{KRC} with $m=2$ is solvable in polynomial time because $\mathcal{X}_2$ contains only two elements. More generally, if $m$ is fixed, \texttt{KRC} remains solvable in polynomial time. This follows from the fact that there are $m!$ potential centroids, and solving \texttt{KRC} requires evaluating subsets of at most $k$ centroids. Once these centroids are fixed, the problem reduces to an assignment problem, which is known to be solvable in polynomial time. While this theoretical polynomial-time complexity is reassuring, it remains computationally prohibitive for practical applications, particularly when $m$ is large. For general values of $m$, \texttt{KRC} is \texttt{NP}-hard, as established in the following theorem, using a reduction from Hypercube Clustering Problem \citep{baldi2012boolean}.

\begin{theorem}[\texttt{NP}-hardness of \texttt{KRC}]
    \label{thm:NPhardness} 
    The \texttt{KRC} problem is \texttt{NP}-hard. 
\end{theorem}

Theorem \ref{thm:NPhardness} demonstrates that solving moderate to large instances of \texttt{KRC} is likely to be computationally intractable in practice, highlighting the need for effective approximation algorithms. To address this challenge, we first establish the analytical solution for the single-cluster special case of \texttt{KRC} in Section \ref{subsec:single_cluster} and derive theoretical error bounds for relaxing \texttt{KRC} to \texttt{KMC} in Section \ref{subsec:relaxation_error_bound}. These results are then leveraged to develop an efficient approximation algorithm for the general case in Section \ref{sec:krca}.


\section{\texttt{KRC} Solutions: Exact for Single-Cluster and Theoretical Bounds for Multiple-Cluster}
\label{sec:single-cluster_error_bounds}

In this section, we establish two key theoretical aspects of the \texttt{KRC} problem. In the single-cluster case, discussed in Section \ref{subsec:single_cluster}, we derive a closed-form solution that can be computed in linear time based on the problem parameters. This result not only has standalone significance but also lays the groundwork for designing approximation methods. For the multi-cluster case, studied in Section \ref{subsec:relaxation_error_bound}, we establish theoretical error bounds for approximating \texttt{KRC} using \texttt{KMC}. Specifically, we show that if a solution to \texttt{KMC} is used as an approximation for \texttt{KRC}, the relative error can exceed 100\%. These findings illustrate the fundamental differences between \texttt{KMC} and \texttt{KRC}, motivating the need for tailored algorithms, which will be presented in Section \ref{sec:krca}.

\subsection{Single-Cluster \texttt{KRC}: Closed-Form Solution}
\label{subsec:single_cluster}

We develop a closed-form solution for the single-cluster special case of \texttt{KRC}, which can be computed in linear time with respect to the problem parameters. This special case holds dual significance: it has standalone applications and serves as a critical component of the approximation algorithm introduced in Section \ref{sec:krca}. Broadly, a clustering algorithm produces two main outputs: the cluster sets and the centroids of the clusters. As discussed earlier, centroids in \texttt{KMC} are determined by computing the mean of the given observation points in each cluster. However, the discrete nature of ranking vectors introduces additional complexity in determining optimal centroids for \texttt{KRC}. 

Given the independence of centroids across clusters, the problem of finding optimal centroids for a set of clusters naturally decomposes into subproblems, each corresponding to a single cluster. Therefore, without loss of generality, we assume a single cluster containing $n$ ranking vectors over $m$ options, with $k=1$ cluster, for ease of presentation.

One approach to solving the single-cluster problem is to formulate it as an integer program and solve it using {commercial solvers}. However, such methods are impractical for large-scale instances due to their prohibitive computation times. Alternatively, the problem can be framed as a ``Minimum Weighted Bipartite Matching'' problem, where the cost of an edge $(j_1, j_2)$ between two nodes $j_1, j_2 \in \{1, \ldots, m\}$ is defined as 
\begin{align}
c_{j_1 j_2} = \sum_{i=1}^n (x_{i j_1} - j_2)^2,
\end{align}
with $x_{i j_1}$ representing the $j_1^{\text{th}}$ component of the ranking vector $\xx_i$. This problem can be solved using the improved Hungarian algorithm by \citet{edmonds1972theoretical} and \citet{tomizawa1971some}, with a worst-case time complexity of $\mathcal{O}(nm^2 + m^3)$. The first term corresponds to the computation of edge costs, and the second term pertains to solving the matching problem. 

Remarkably, we present a significantly faster procedure for solving the single-cluster case of \texttt{KRC}, as provided in Theorem \ref{thm:optimal_centeriod}. This theorem offers an analytical closed-form solution for determining the optimal centroid, which can be computed in linear time $\mathcal{O}(nm)$ with respect to the number of observations $n$ and options $m$.

\begin{theorem}[Optimal Centroid]
    \label{thm:optimal_centeriod}
    Consider a single cluster $\mathcal{S} = \{\xx_1, \ldots, \xx_n \}$, {where 
    $\xx_i= \begin{bmatrix} x_{i1} & \cdots & x_{im} \end{bmatrix} \in \mathcal{X}_m$
    represents} the $i^{\text{th}}$ ranking vector for $i = 1, \ldots, n$. Define
    \begin{align*}
        \bar{x}_{.j} & = \frac{\sum_{i=1}^{n} x_{ij}}{n} \quad \text{for } j = 1, \ldots, m,
    \end{align*}
    and sort these values in non-decreasing order. If $y_j^*$ is defined as the rank of $\bar{x}_{.j}$ in this sorted sequence, {then 
    $\yy^* = \begin{bmatrix} y_1^* & \cdots & y_m^* \end{bmatrix} \in \mathcal{X}_m$
    is} an optimal centroid for the cluster. Additionally, if there are ties among $\bar{x}_{.j}$ values, any arbitrary ordering of the ties will also be optimal.
\end{theorem}

Consequently, similar to the \texttt{KMC} problem, we simplify the notation of the objective function \eqref{eq:objFn_krc} by omitting the explicit representation of centroids $\yy_1, \ldots, \yy_k$. The objective function is denoted as $v_{\mathtt{KRC}}(\mathcal{S}_1, \ldots, \mathcal{S}_k)$, implying that centroids are set optimally {according to Theorem \ref{thm:optimal_centeriod}.}

Theorem \ref{thm:optimal_centeriod} highlights that the primary computational challenge of the \texttt{KRC} problem lies in determining the optimal clusters. Consequently, developing an efficient approximation algorithm for the \texttt{KRC} problem reduces to devising a method for approximately clustering the observed ranking vectors. Since \(\mathcal{X}_m \subset \mathbb{R}^m\), the \texttt{KMC} problem can be regarded as a ``relaxation'' of the \texttt{KRC} problem. Thus, it may be natural to propose using a solution to the \texttt{KMC} problem as an approximation for the \texttt{KRC} problem. Specifically, one can compute a locally optimal clustering for the \texttt{KMC} problem, use it as an initial approximation for the \texttt{KRC} problem, and then determine the optimal centroids by applying Theorem \ref{thm:optimal_centeriod}. 

However, {as we demonstrate theoretically in Section \ref{subsec:relaxation_error_bound} and numerically} in Section \ref{sec:numerical_experiments}, such a relaxation can result in significant relative errors when approximating the \texttt{KRC} problem. In some cases, these errors may exceed $100\%$, highlighting the limitations of directly applying this approach.

\subsection{Relaxing \texttt{KRC} via \texttt{KMC}: Theoretical Bounds and Errors} 
\label{subsec:relaxation_error_bound}

As discussed earlier, $\mathcal{X}_m \subset \mathbb{R}^m$ implies that \texttt{KMC} can be viewed as a relaxation of \texttt{KRC}, meaning that $v_{\mathtt{KMC}}^* \leq v_{\mathtt{KRC}}^*$. Consequently, the optimal value of \texttt{KMC} serves as a lower bound for the optimal value of \texttt{KRC}. Next, we establish an upper bound for \texttt{KRC} in Proposition \ref{pro:upper_bnd} and Corollary \ref{cor:upper_bnd}.

\begin{proposition}[Upper Bound for General Clustering]
    \label{pro:upper_bnd}
    Consider the given clusters $\mathcal{S}_1, \ldots, \mathcal{S}_k$ for $n$ ranking vectors from $\mathcal{X}_m$. We have
    \begin{align*}
        v_{\mathtt{KRC}}(\mathcal{S}_1, \ldots, \mathcal{S}_k) & \leq 2 v_{\mathtt{KMC}}(\mathcal{S}_1, \ldots, \mathcal{S}_k).
    \end{align*}
\end{proposition}

\begin{corollary}[Upper and Lower Bounds for Optimal Clustering]
    \label{cor:upper_bnd}
    Let $\mathcal{S} = \{\xx_1, \ldots, \xx_n\}$, where $\xx_i \in \mathcal{X}_m$ for $i=1, \ldots, n$. Given $\mathcal{S}$, let $v_{\mathtt{KMC}}^*$ and $v_{\mathtt{KRC}}^*$ denote the optimal values of \texttt{KMC} and \texttt{KRC}, respectively. We have 
    \begin{align*}
        v_{\mathtt{KMC}}^* & \leq v_{\mathtt{KRC}}^* \leq 2 v_{\mathtt{KMC}}^*.
    \end{align*}
\end{corollary}

Corollary \ref{cor:upper_bnd} states that if we relax \texttt{KRC} and instead solve \texttt{KMC} as an approximate clustering for the former problem, the relative error for this approximation could be as large as $50\%$, that is,
\begin{align*}
    \frac{v_{\mathtt{KRC}}^* - v_{\mathtt{KMC}}^*}{v_{\mathtt{KRC}}^*} & \leq \frac{1}{2}.
\end{align*}
Next, we demonstrate the tightness of this bound by constructing an instance of \texttt{KRC} that achieves the above worst-case bound.

\subsubsection{Tightness of the Error Bound.} 
\label{subsubsec:tightness_error}
    We first consider the simplest case of $k=1$ with two observations, $\xx_1 = \begin{bmatrix} 1 & 2 \end{bmatrix}$ and $\xx_2 = \begin{bmatrix} 2 & 1 \end{bmatrix}$. It is readily observed that the optimal centroids of the \texttt{KRC} and \texttt{KMC} problems are $\yy_{\texttt{KRC}}^* = \begin{bmatrix} 1 & 2 \end{bmatrix}$ and $\yy_{\texttt{KMC}}^* = \begin{bmatrix} 1.5 & 1.5 \end{bmatrix}$, respectively (note that \texttt{KRC} has an alternative optimal centroid $\yy_{\texttt{KRC}}^* = \begin{bmatrix} 2 & 1 \end{bmatrix}$). Hence, $v_{\texttt{KRC}}^* = 2$ and $v_{\texttt{KMC}}^* = 1$, achieving the upper bound in Corollary \ref{cor:upper_bnd}.
    
    Next, we construct an example for $k=2$ with four observations $\xx_1=\begin{bmatrix} 1 & 2 & 3 & 4 & 5 & 6 & 7 \end{bmatrix}$, $\xx_2=\begin{bmatrix} 2 & 1 & 3 & 4 & 5 & 6 & 7 \end{bmatrix}$, $\xx_3=\begin{bmatrix} 2 & 1 & 5 & 4 & 3 & 6 & 7 \end{bmatrix}$, and $\xx_4=\begin{bmatrix} 2 & 1 & 5 & 4 & 3 & 7 & 6 \end{bmatrix}$. These four observations are constructed such that the two pairs $(\xx_1, \xx_2)$ and $(\xx_3, \xx_4)$ have a squared Euclidean distance of two, and any other pair of observations has a squared Euclidean distance sufficiently greater than two. One can easily verify that $\mathcal{S}_1 = \{\xx_1, \xx_2\}$ and $\mathcal{S}_2 = \{\xx_3, \xx_4\}$ are optimal clusters for both the \texttt{KRC} and \texttt{KMC} problems, with $v_{\texttt{KRC}}^* = 4$ and $v_{\texttt{KMC}}^* = 2$, respectively, again achieving the given upper bound.
    
    Motivated by these two numerical examples, we generalize this idea to any number of clusters $k$, where $k \in \{1, 2, \dots\}$, by constructing instances in which $v_{\mathtt{KRC}}^* = 2 v_{\mathtt{KMC}}^*$. For a given value of $k$, let $\bar{m}$ denote the smallest $\eta \in \mathbb{Z}^+$ that satisfies 
    \begin{align*}
        \left\| \begin{bmatrix} \eta & \eta-1 & \cdots & 1 \end{bmatrix} - \begin{bmatrix} 1 & 2 & \cdots & \eta \end{bmatrix} \right\|^2 & \geq 4k,
    \end{align*}
    or equivalently, 
    \begin{align*}
        \frac{(\eta-1) \eta (\eta+1)}{3} & \geq 4k.
    \end{align*}
    Clearly, such an $\bar{m}$ exists and $\bar{m} \geq 3$. Now, let $n = 2k$, $m = \bar{m}(k-1) + 2k$, and define 
    \begin{align*} 
    \theta_i =
      \begin{cases}
        (\bar{m}+2)(\frac{i-1}{2}-1)+2 & \text{for } i = 1,3,\ldots, 2k-1,\\
        (\bar{m}+2)(\frac{i}{2}-1) & \text{for } i = 2,4,\ldots, 2k,
      \end{cases}
    \end{align*}
    and 
    \begin{align*} 
    \lambda_i =
      \begin{cases}
        \bar{m} & \text{for } i = 1,3,\ldots, 2k-1,\\
        2 & \text{for } i = 2,4,\ldots, 2k.
      \end{cases}
    \end{align*}
    \noindent Construct the set of $n$ observations by setting $\hat{\xx}_1 = \begin{bmatrix} 1 & \cdots & m \end{bmatrix}$ and forming the observations $\hat{\xx}_i = \begin{bmatrix} \hat{x}_{i,1} & \cdots & \hat{x}_{i,m} \end{bmatrix}$ for $i=2,\ldots,n$ as follows:
    \begin{align*} 
    \hat{x}_{i,j} & =
      \begin{cases}
        \hat{x}_{i-1,j} & \text{for } j \leq \theta_i,\\
        \lambda_i + 2\theta_i - j + 1 & \text{for } \theta_i + 1 \leq j \leq \theta_i + \lambda_i,\\
        j & \text{for } j \geq \theta_i + \lambda_i + 1.
      \end{cases}
    \end{align*}
    \noindent One can show that for any pair of indices $1 \leq i_1 < i_2 \leq n$, if $i_1$ is an odd number and $i_2 = i_1 + 1$, we have $\| \xx_{i_2} - \xx_{i_1} \|^2 = 2$; otherwise, $\| \xx_{i_2} - \xx_{i_1} \|^2 \geq 4k$. 
    
    We next argue that the clusters $\hat{\mathcal{S}}_i = \{\xx_{2i-1}, \xx_{2i}\}$ with centroid $\xx_{2i-1}$ for $i=1,\ldots,k$, form an optimal solution for \texttt{KRC}. Clearly, this is a feasible solution with the objective value of $v_{\mathtt{KRC}}(\hat{\mathcal{S}}_1, \dots, \hat{\mathcal{S}}_k) = 2k$. It can be readily observed that for any two distinct observations $\xx_1, \xx_2 \in \mathcal{X}_m$, we have $\| \xx_2 - \xx_1 \|^2 \geq 2$. Furthermore, for any given $k$ clusters with $2k$ observations in $\mathcal{X}_m$, as the $k$ centroids also belong to $\mathcal{X}_m$, calculating the objective function $v_{\texttt{KRC}}$ requires the summation of at least $k$ such difference squared norms, which implies that it is bounded below by $2k$. Thus, this demonstrates the optimality of the clusters $\hat{\mathcal{S}}_1, \dots, \hat{\mathcal{S}}_k$ for \texttt{KRC}, that is, $v_{\texttt{KRC}}^* = 2k$. 
    
    Lastly, consider the \texttt{KMC} problem for the constructed instance. The clusters $\hat{\mathcal{S}}_1, \dots, \hat{\mathcal{S}}_k$ with centroids $\frac{1}{2}(\xx_1+\xx_2)$, ..., $\frac{1}{2}(\xx_{n-1}+\xx_{n})$, respectively, form a feasible solution for \texttt{KMC}, achieving the objective value of $v_{\texttt{KMC}}(\hat{\mathcal{S}}_1, \dots, \hat{\mathcal{S}}_k) = k$. Additionally, Corollary \ref{cor:upper_bnd} states that $v_{\texttt{KMC}}^* \geq \frac{1}{2}v_{\texttt{KRC}}^* = k$. Therefore, the clusters $\hat{\mathcal{S}}_1, \dots, \hat{\mathcal{S}}_k$ achieve the lower bound of the corresponding objective value, implying that they are optimal, that is, $v_{\texttt{KMC}}^* = k$. Hence, $v_{\texttt{KRC}}^* = 2 v_{\texttt{KMC}}^*$.

\subsubsection{Theoretical Error Bound.}
\label{subsubsec:theoretical_error}
The relative error provided by Proposition \ref{pro:upper_bnd} applies to the case when \texttt{KMC} is solved optimally. However, due to its \texttt{NP}-hardness, solving real-size instances of \texttt{KMC} optimally is not tractable. Thus, as discussed in Section \ref{subsec:lit_theory&algo}, several approximation algorithms have been proposed in the literature to guarantee finding an $\varepsilon$-optimal solution for \texttt{KMC}, for some $\varepsilon > 0$, meaning that the objective value of the solution is no larger than $(1+\varepsilon)$ times the optimal value. 

More precisely, if $\bar{\mathcal{S}}_1, \ldots, \bar{\mathcal{S}}_k$ is a solution produced by such an approximation algorithm, we have
\begin{align}\label{eq:eps-optimal_kmc}
    v_{\mathtt{KMC}}(\bar{\mathcal{S}}_1, \ldots, \bar{\mathcal{S}}_k) & \leq (1+\varepsilon) v_{\mathtt{KMC}}^*.
\end{align}
Such an approximation can also be applied to solve \texttt{KRC}. The resulting solution is a feasible clustering for \texttt{KRC}, and one needs to apply Theorem \ref{thm:optimal_centeriod} to compute an optimal centroid for each cluster. Notably, using Inequality (\ref{eq:eps-optimal_kmc}), Proposition \ref{pro:upper_bnd}, and Corollary \ref{cor:upper_bnd}, we establish in Proposition \ref{pro:eps-optimal_krc} that this approach results in a larger relative error of $(1+2\varepsilon)$ in the worst case.

\begin{proposition}[$(1+2\varepsilon)$-Optimal Approximation]
    \label{pro:eps-optimal_krc}
    Let $\mathcal{S} = \{\xx_1, \ldots, \xx_n\}$, where $\xx_i \in \mathcal{X}_m$ for $i=1, \ldots, n$. 
    Let $\bar{\mathcal{S}}_1, \ldots, \bar{\mathcal{S}}_k$ be an $\varepsilon$-optimal solution for \texttt{KMC}. This clustering serves as a $(1+2\varepsilon)$-optimal solution for \texttt{KRC}, that is,
    \begin{align*}
        v_{\mathtt{KRC}}(\bar{\mathcal{S}}_1, \ldots, \bar{\mathcal{S}}_k) & \leq (2+2\varepsilon) v_{\mathtt{KRC}}^*,
    \end{align*}    
    or equivalently, assuming $v_{\mathtt{KRC}}^* \neq 0$,
    \begin{align*}
        \frac{v_{\mathtt{KRC}}(\bar{\mathcal{S}}_1, \ldots, \bar{\mathcal{S}}_k) - v_{\mathtt{KRC}}^*}{v_{\mathtt{KRC}}^*} & \leq 1+2\varepsilon.
    \end{align*}
\end{proposition}

Proposition \ref{pro:eps-optimal_krc} states that while an $\varepsilon$-optimal solution for the \texttt{KMC} problem can serve as a reasonable approximation for \texttt{KMC}, applying this approximation to the \texttt{KRC} problem may result in a significant error, potentially exceeding $100\%$. Through extensive numerical experiments on both real and synthetic data (Section \ref{sec:numerical_experiments}), we demonstrate that this considerable relative error is not merely a by-product of theoretical analysis but can indeed occur in practice.

Furthermore, we {intuitively} illustrate that an optimal clustering for the \texttt{KMC} problem is not necessarily optimal for the \texttt{KRC} problem. The intuition behind this observation lies in the distinct nature of centroids in the two problems. While centroids in \texttt{KMC} are simply the averages of the points in each cluster, centroids in \texttt{KRC} are restricted to be ranking vectors from the set $\mathcal{X}_m$. In other words, the centroid of a cluster in \texttt{KRC} shifts from the real-valued mean to a nearby discrete point in $\mathcal{X}_m$. Consequently, a point near the boundary of a cluster in \texttt{KMC} may be better assigned to a different cluster, leading to a different optimal clustering for \texttt{KRC}.

These theoretical and numerical findings highlight the critical need to develop an efficient approximation algorithm that captures the specific structure of the \texttt{KRC} problem and provides a more suitable approximation. {We proceed to develop our new approximation algorithm.} 


\section{\texttt{KRCA}: Efficient Approximation Algorithm for \texttt{KRC}}
\label{sec:krca}

In this section, we introduce an efficient approximation algorithm, \texttt{KRCA}, and establish its theoretical properties. Algorithm \ref{alg:generic_framework} outlines the general framework of \texttt{KRCA}, which comprises three main steps: initialization, updating centroids, and reconstructing clusters. Although \texttt{KRCA} shares structural similarities with Lloyd's algorithm for \texttt{KMC} \citep{lloyd1982least}, each step is specifically tailored to exploit the unique characteristics of ranking vectors.

Initially, the algorithm creates a set of clusters for the given ranking vectors. It then iteratively updates the centroids and reassigns each ranking vector to its closest centroid. {To enhance efficiency, for the instances with small $m$,} we introduce a novel branch-and-bound algorithm, \texttt{BnB}, for cluster reconstruction (Algorithm \ref{alg:b&b}). This approach outperforms an {exhaustive search (\texttt{ES}) method,} which compares each ranking vector with all updated centroids to find the closest one, {for small values of $m$.} However, when $m$ becomes large, \texttt{BnB} may face computational challenges. {To reduce the computational time of \texttt{BnB},} we introduce a hyperparameter $\epsilon \geq 0$ that controls the trade-off between accuracy and computational efficiency. By adjusting $\epsilon$, we can significantly speed up \texttt{BnB} at the cost of generating approximate solutions. Notably, when $\epsilon = 0$, \texttt{BnB} produces the exact clustering solution obtained by the {\texttt{ES} method}. More generally, the inclusion of $\epsilon$ enhances \texttt{BnB}'s efficiency {for larger values of $m$}. The iterative process (lines 4 and 5 of Algorithm \ref{alg:generic_framework}) continues until predefined stopping criteria, described below, are satisfied.

\begin{algorithm}[h!]
    \OneAndAHalfSpacedXI\small\footnotesize
    \begin{algorithmic}[1]
        \caption{\small Generic Framework of the \texttt{KRCA} Algorithm
        \label{alg:generic_framework}}
        \Begin
        \State \textsc{Inputs:} The set of ranking vectors and the number of clusters
        \State Initialization
        \While{Stopping criteria not met}
            \State Updating centroids
            \State Reconstructing clusters
        \EndWhile
        \State \textsc{Outputs:} A set of clusters with their optimal centroids
        \End
    \end{algorithmic}
\end{algorithm}

\begin{enumerate}[label=\ \ \ \ \ (\roman*)\ \ , align=left, leftmargin=0pt, labelsep=0em, itemsep=0ex, listparindent=0pt]
    \item \textbf{Initialization. }As highlighted in Section \ref{subsec:relaxation_error_bound}, approximation algorithms for \texttt{KMC} can generate $\varepsilon$-optimal solutions, which provide a strong foundation as initial clusters for \texttt{KRC}. Proposition \ref{pro:eps-optimal_krc} guarantees that this initial approximation already yields a $(1+2\varepsilon)$-optimal solution for \texttt{KRC}, as each iteration of the algorithm is designed to maintain or reduce the objective value. By leveraging these advancements, our algorithm not only builds upon existing \texttt{KMC} approximation methods but also establishes a robust starting point for \texttt{KRCA}.

    \item \textbf{Updating Centroids. }In this step, the set of clusters along with their corresponding points is provided, and the optimal centroids need to be determined. Since the computation of the optimal centroid for each cluster is independent of the others, Theorem \ref{thm:optimal_centeriod} is applied to efficiently calculate the optimal centroid for each cluster.
    
    \item \textbf{Reconstructing Clusters. }In this step, the set of updated centroids is provided, and each observation ranking vector is assigned to its closest centroid. An {\texttt{ES} method}, which involves comparing the Euclidean distance between each of the $n$ observation ranking vectors and $k$ centroids, requires $\bigO{nmk}$ operations. {We propose a novel branch-and-bound algorithm, \texttt{BnB}, to enhance the computational performance of reconstructing clusters for small values of $m$. 
    %
    %
    The \texttt{BnB} algorithm uses a decision tree framework to efficiently assign ranking vectors to centroids (Algorithm \ref{alg:b&b}). It partitions the ranking vectors into subsets using branching, with each subset represented as a node in the tree.} 
    
    {The \texttt{KRCA} algorithm applies \texttt{BnB} when $m$ is small and switches to the \texttt{ES} method when $m$ exceeds a threshold, thereby exploiting the advantages of both algorithms. The threshold can be learned, as demonstrated in Section~\ref{subsec:synthetic_data}.} 
    The \texttt{BnB} algorithm offers the following key advantages.
        \begin{enumerate}
            \item \textit{Improved computation time for small-$m$ instances}: \texttt{BnB} efficiently processes cases where $m$ is small and $n$ is large, reducing the computational burden compared to the {\texttt{ES} method} {(Section~\ref{subsec:synthetic_data})}.
            \item \textit{Error-control mechanism}: A hyperparameter $\epsilon$ enables \texttt{BnB} to balance solution quality and computational efficiency, allowing for suboptimal solutions with bounded errors. This flexibility extends its applicability to {larger $m$ values (Theorem~\ref{thm:alg_b&b_error} and Section~\ref{subsubsec:effects_of_controlling_param}).}
            \item \textit{Exploitation of input structure}: \texttt{BnB} leverages the inherent clustering within input ranking vectors, further reducing computation time for well-structured datasets {(Theorem~\ref{thm:tree_size}  and Section~\ref{subsubsec:clustered_rv}).}
        \end{enumerate}
        
        In Section \ref{subsec:analytical_properties_b&b}, we provide a detailed explanation of \texttt{BnB}, develop its analytical properties, and explore the characteristics of the solutions it produces.
        
        \begin{algorithm}[ht!]
            \OneAndAHalfSpacedXI\small\footnotesize
            \begin{algorithmic}[1]
                \caption{\small Branch-and-Bound \texttt{BnB} Algorithm for Cluster Reconstruction\label{alg:b&b}}
                \Begin
                \State \textsc{Inputs:} The set of observation ranking vectors $\mathcal{S}$, the set of centroids $\mathcal{C}$, and the controlling parameter $\epsilon \in \mathbb{R}^+$
                \State Initialize $\Phi=\{\phi_{0}\}$, $\mathcal{S}_{\phi_{0}} = \mathcal{S}$, $\ff_{\phi_{0}} = \begin{bmatrix}
                \ \end{bmatrix}$, $m_{\phi_{0}} = 0$, $\mathcal{C}_{\phi_{0}} = \mathcal{C}$, $c=0$
                \While{$\Phi\ne\{\}$}
                \State Select an arbitrary node $\phi \in \Phi$ and remove it
                \For{($j \in \{1, \dots, m\}$) \textbf{and} ($j$ not fixed in $\ff_{\phi}$)} 
                \If{(there exists $\xx \in \mathcal{S}_{\phi}$ such that its $(m_{\phi} + 1)^{\text{st}}$ coordinate, $x_{m_{\phi} + 1}$, equals $j$)}
                \State Increment $c = c + 1$
                \State Create child node $\phi_{c}$ with:
                \State \quad \quad $\ff_{\phi_c} = [\ff_{\phi}\ j]$, $m_{\phi_c} = m_\phi + 1$
                \State \quad \quad $\mathcal{S}_{\phi_c} = \{\xx \in \mathcal{S}_{\phi} \mid x_{m_{\phi_c}} = j\}$
                \State \quad \quad $\mathcal{C}_{\phi_c} = \mathcal{C}_{\phi}$
                \State \quad \quad Compute the lower and upper bounds $\mathtt{LB}_{\phi_{c},\yy}$ and $\mathtt{UB}_{\phi_{c},\yy}$ for all $\yy \in \mathcal{C}_{\phi_{c}}$:
                \State \quad \quad \quad \quad $\mathtt{LB}_{\phi_c,\yy} = \| \yy(1:m_{\phi_c}) - \ff_{\phi_c} \|^2 + \| \underline{\yy}(m_{\phi_c}+1:m) - \underline{\ff}_{\phi_c} \|^2$
                \State \quad \quad \quad \quad $\mathtt{UB}_{\phi_c,\yy} = \| \yy(1:m_{\phi_c}) - \ff_{\phi_c} \|^2 + {\| \underline{\yy}(m_{\phi_c}+1:m) - \overline{\ff}_{\phi_c} \|^2}$
                \State \quad \quad where $\underline{\yy}(m_{\phi_c}+1:m)$ denotes the ascending orderings of the vector $\yy(m_{\phi_c}+1:m)$. 
                \State \quad \quad Similarly, $\underline{\ff}_{\phi_c}$ and $\overline{\ff}_{\phi_c}$ are vectors of size $m - m_{\phi_c}$ containing all integers in $\{1, \dots, m\} \setminus \ff_{\phi_c}$, 
                \State \quad \quad sorted in ascending and descending order, respectively.
                \For{($\yy\in\mathcal{C}_{\phi_c}$)} 
                \If{($\mathtt{LB}_{\phi_c,\yy} \ge \min_{\yy'\in\mathcal{C}_{\phi_c}\backslash\{\yy\}} \mathtt{UB}_{\phi_c,\yy'}-\epsilon$)}
                \State Remove $\yy$ from $\mathcal{C}_{\phi_c}$
                \EndIf
                \EndFor
                \If{($|\mathcal{C}_{\phi_c}|=1$)}
                \State Assign all vectors in $\mathcal{S}_{\phi_c}$ to the singleton centroid in $\mathcal{C}_{\phi_c}$
                \Else
                \State Add $\phi_c$ to $\Phi$
                \EndIf
                \EndIf
                \EndFor
                \EndWhile
                \State \textsc{Outputs:} Clusters with centroids in $\mathcal{C}$
                \End
            \end{algorithmic}
        \end{algorithm}
    
    \item \textbf{Stopping criteria. }Several stopping criteria can be utilized, such as terminating the algorithm when the absolute difference between the objective values in two consecutive iterations falls below a specified threshold, when the number of iterations reaches a predefined limit, or a combination of these conditions.
\end{enumerate}

\subsection{\texttt{BnB}: Analytical Properties and Computational Advantages} 
\label{subsec:analytical_properties_b&b}

In this section, we elaborate on the \texttt{BnB} algorithm (Algorithm \ref{alg:b&b}) for reconstructing clusters during iterations of \texttt{KRCA} and establish its theoretical convergence properties. As shown in Algorithm \ref{alg:b&b}, the inputs for \texttt{BnB} include the set of observation ranking vectors $\mathcal{S}$, the set of centroids $\mathcal{C}$, and a controlling hyperparameter $\epsilon \in \mathbb{R}^{+}$ that allows early pruning of branches (Line 1 of Algorithm \ref{alg:b&b}). The algorithm outputs a set of $k$ clusters, where each observation ranking vector is assigned to the closest centroid, with an error controlled by the parameter $\epsilon$ (Line 31 of Algorithm \ref{alg:b&b}).

\subsubsection{{The Iterative Process in \texttt{BnB}.}} 
\label{subsubsec:illustration_b&b}
\begin{figure}[h!]
        \centering
        \includegraphics[width=1\textwidth,trim=3.1cm 12.5cm 12.9cm 3.2cm]{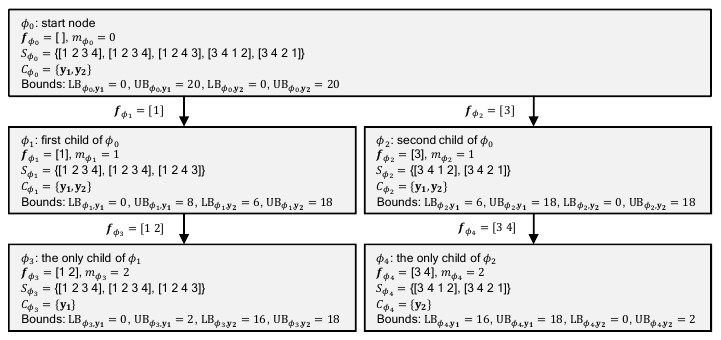}
        \caption{An illustrative example of the \texttt{BnB} algorithm (Algorithm \ref{alg:b&b}) with parameters $n=5$ observations, $m=4$ options, $k=2$ clusters, and the controlling parameter $\epsilon = 0$. The set of observation ranking vectors is given by $\mathcal{S} = \left\{ \begin{bmatrix} 1 & 2 & 3 & 4 \end{bmatrix}, \begin{bmatrix} 1 & 2 & 3 & 4 \end{bmatrix}, \begin{bmatrix} 1 & 2 & 4 & 3 \end{bmatrix}, \begin{bmatrix} 3 & 4 & 1 & 2 \end{bmatrix}, \begin{bmatrix} 3 & 4 & 2 & 1 \end{bmatrix} \right\}$, and the set of centroids is $\mathcal{C} = \{\yy_1, \yy_2\}$, where $\yy_1 = \begin{bmatrix} 1 & 2 & 3 & 4 \end{bmatrix}$ and $\yy_2 = \begin{bmatrix} 3 & 4 & 1 & 2 \end{bmatrix}$.
        }
        \label{fig:bnb_example}
    \end{figure}

To illustrate the process, we use a simple example depicted in Figure~\ref{fig:bnb_example}. In this example, there are $n=5$ observation ranking vectors over $m=4$ options, represented as $\mathcal{S} = \left\{ \begin{bmatrix} 1 & 2 & 3 & 4 \end{bmatrix}, \begin{bmatrix} 1 & 2 & 3 & 4 \end{bmatrix}, \begin{bmatrix} 1 & 2 & 4 & 3 \end{bmatrix}, \begin{bmatrix} 3 & 4 & 1 & 2 \end{bmatrix}, \begin{bmatrix} 3 & 4 & 2 & 1 \end{bmatrix} \right\}$. These vectors are to be partitioned into $k=2$ clusters with the centroids $\mathcal{C} = \{\yy_1, \yy_2 \}$, where $\yy_1 = \begin{bmatrix} 1 & 2 & 3 & 4 \end{bmatrix}$ and $\yy_2 = \begin{bmatrix} 3 & 4 & 1 & 2 \end{bmatrix}$. The controlling parameter is set to $\epsilon = 0$, ensuring optimal clustering such that each ranking vector is assigned to the closest centroid.
    
    
    
    Let $\Phi$ denote the set of active nodes in the tree, represented by the boxes in Figure~\ref{fig:bnb_example}. This set is updated as Algorithm~\ref{alg:b&b} iterates. For each node $\phi \in \Phi$, the following notations are defined:
    \begin{itemize}
        \item $\mathcal{S}_{\phi} \subseteq \mathcal{S}$: the set of observation ranking vectors belonging to node $\phi$,
        \item $\ff_\phi$: a tuple fixing the first entries of all observation ranking vectors in $\mathcal{S}_{\phi}$,
        \item $m_\phi$: the size of the tuple $\ff_\phi$,
        \item $\mathcal{C}_\phi \subseteq \mathcal{C}$: the set of available centroids at node $\phi$.
    \end{itemize}
    
    Initially, $\Phi$ contains a single element, the initial node $\phi_{0}$, which includes all observation ranking vectors. Thus, $\mathcal{S}_{\phi_{0}} = \mathcal{S}$, $\ff_{\phi_{0}} = \begin{bmatrix} \ \end{bmatrix}$, $m_{\phi_0} = 0$, and $\mathcal{C}_{\phi_0} = \mathcal{C}$ (Line 2 of Algorithm~\ref{alg:b&b}). 
    
    In the outer while-loop (Lines 3-30 of Algorithm~\ref{alg:b&b}), the only element in the active set $\Phi$, namely $\phi_0$, is selected and removed. Then, in the inner for-loop (Lines 5-29 of Algorithm~\ref{alg:b&b}), all possible child nodes are constructed by fixing the first element of observation ranking vectors in each node based on $\mathcal{S}_{\phi_0}$. In Figure~\ref{fig:bnb_example}, based on the input ranking vectors in $\phi_0$, two possible child nodes are generated: $\phi_1$ and $\phi_2$. These nodes have $\ff_{\phi_1} = \begin{bmatrix} 1 \end{bmatrix}$ and $\ff_{\phi_2} = \begin{bmatrix} 3 \end{bmatrix}$, respectively, with $m_{\phi_1} = 1$ and $m_{\phi_2} = 1$. Specifically, child node $\phi_1$ includes all observation ranking vectors starting with $1$, and child node $\phi_2$ includes all observation ranking vectors starting with $3$. 
    
    The lower and upper bounds $\mathtt{LB}_{\phi_{c},\yy}$ and $\mathtt{UB}_{\phi_{c},\yy}$ for each centroid in node $\phi_{c}$ for $c = 1, 2$ are calculated as defined in Lines 13-14 of Algorithm~\ref{alg:b&b}. To compute these bounds, we first determine $\underline{\ff}_{\phi_c}$ and $\overline{\ff}_{\phi_c}$ for each node $\phi_c$ as follows:
    \begin{itemize}
        \item For $c = 1$: $\underline{\ff}_{\phi_1} = \begin{bmatrix} 2 & 3 & 4 \end{bmatrix}$ and $\overline{\ff}_{\phi_1} = \begin{bmatrix} 4 & 3 & 2 \end{bmatrix}$.
        \item For $c = 2$: {$\underline{\ff}_{\phi_2} = \begin{bmatrix} 1 & 2 & 4 \end{bmatrix}$ and $\overline{\ff}_{\phi_2} = \begin{bmatrix} 4 & 2 & 1 \end{bmatrix}$.}
    \end{itemize}
    These bounds help eliminate centroids that are unlikely to be optimal, allowing the \texttt{BnB} algorithm to focus computational efforts on the most promising nodes. Nodes are pruned when they either contain no ranking vectors or have only one viable centroid.
   
    Since no centroids are eliminated at either child node, $\phi_1$ and $\phi_2$, the set of active nodes is updated to $\Phi = \{\phi_1, \phi_2\}$. Next, one of the nodes in $\Phi$, say $\phi_1$, is selected and removed. This node generates a single child, $\phi_3$, with $\ff_{\phi_3} = \begin{bmatrix} 1 & 2 \end{bmatrix}$, $m_{\phi_3} = 2$, and $\mathcal{C}_{\phi_3} = \{\yy_1, \yy_2\}$. After computing the bounds for $\yy_1$ and $\yy_2$, it is determined that $\mathtt{LB}_{\phi_3,\yy_2} \geq \mathtt{UB}_{\phi_3,\yy_1}$, leading to the removal of $\yy_2$ from $\mathcal{C}_{\phi_3}$. Consequently, $\mathcal{C}_{\phi_3} = \{\yy_1\}$ becomes a singleton, and node $\phi_3$ is not added to the set of active nodes, which is then updated to $\Phi = \{\phi_2\}$. 
    
    Following a similar process, node $\phi_2$ generates a child $\phi_4$, where $\mathcal{C}_{\phi_4}$ also reduces to a singleton. At this stage, the set of active nodes becomes empty, and the while-loop terminates. The algorithm produces two clusters, $\mathcal{S}_{\phi_3}$ and $\mathcal{S}_{\phi_4}$, with centroids $\yy_1$ and $\yy_2$, respectively.

\subsubsection{Convergence Properties of \texttt{BnB}.}
\label{subsubsec:convergence_properties_bnb}
Two critical questions concerning the \texttt{BnB} algorithm pertain to its convergence and computational efficiency. These are addressed in Theorem \ref{thm:alg_b&b_error} and Theorem \ref{thm:tree_size} below. 
\begin{theorem}[Theoretical Error Bound for the \texttt{BnB} Algorithm]
    \label{thm:alg_b&b_error}
    Consider the \texttt{KRC} problem with $n$ observation ranking vectors and $k$ clusters, and given centroids $\yy_1, \ldots, \yy_k$ at a cluster reconstruction step of the \texttt{BnB} algorithm. Let 
    \begin{align*}
        v_{\mathtt{KRC}}(\mathcal{S}_1^*, \ldots, \mathcal{S}_k^*, \yy_1, \ldots, \yy_k) & = \min_{\mathcal{S}_1, \ldots, \mathcal{S}_k}\sum_{\ell=1}^{k}\sum_{\xx \in \mathcal{S}_{\ell}} \| \xx - \yy_{\ell} \|^2,
    \end{align*}
    be the optimal objective value with the optimal clusters $\mathcal{S}_1^*, \ldots, \mathcal{S}_k^*$, and let the clusters $\bar{\mathcal{S}}_1, \ldots, \bar{\mathcal{S}}_k$ generated by the output of the \texttt{BnB} algorithm for $\epsilon \geq 0$ have the corresponding objective value
    \begin{align*}
        v_{\mathtt{KRC}}(\bar{\mathcal{S}}_1, \ldots, \bar{\mathcal{S}}_k, \yy_1, \ldots, \yy_k) & = \sum_{\ell=1}^{k}\sum_{\xx \in \bar{\mathcal{S}}_{\ell}} \| \xx - \yy_{\ell} \|^2.
    \end{align*}
    Then, we have
    \begin{align*}
         v_{\mathtt{KRC}}(\bar{\mathcal{S}}_1, \ldots, \bar{\mathcal{S}}_k, \yy_1, \ldots, \yy_k) - v(\mathcal{S}_1^*, \ldots, \mathcal{S}_k^*, \yy_1, \ldots, \yy_k) & \leq n(k-1)\epsilon.
    \end{align*}
\end{theorem}

\begin{corollary}[Convergence Guarantee of the \texttt{BnB} Algorithm]
    \label{cor:convergence_bnb}
    The \texttt{BnB} algorithm generates an optimal clustering for $\epsilon = 0$.
\end{corollary}

Theorem \ref{thm:alg_b&b_error} establishes an error bound for the output of Algorithm \ref{alg:b&b} in terms of the input hyperparameter $\epsilon$. Corollary \ref{cor:convergence_bnb} further demonstrates that when $\epsilon = 0$, the algorithm operates without approximation, guaranteeing both the convergence and optimality of \texttt{BnB}.

Allowing $\epsilon > 0$ introduces flexibility to the \texttt{BnB} algorithm by enabling early pruning of nodes (Line 19 of Algorithm \ref{alg:b&b}). This reduces computation time while maintaining control over the trade-off between solution quality and computational efficiency. This adaptability offers a significant advantage over the {\texttt{ES} method}, which lacks such flexibility.

We now discuss the computational advantages of \texttt{BnB}. As illustrated in Figure~\ref{fig:bnb_example}, \texttt{BnB} constructs the first layer of the tree (nodes $\phi_1$ and $\phi_2$) by assigning each ranking vector to a node based on its first entry and computing the lower and upper bounds. This requires approximately $n + d$ operations, where $d$ represents the number of operations needed to compute the bounds in $\phi_1$ and $\phi_2$. Similarly, constructing the second layer (nodes $\phi_3$ and $\phi_4$) requires another $n + d$ operations, leading to a total of approximately $2n + 2d$ operations.

Now, consider a scaled version of the example in Figure~\ref{fig:bnb_example}, where $m$, $k$, $\yy_1$, $\yy_2$, and $\epsilon$ remain unchanged, but the set of observation ranking vectors $\mathcal{S}$ is expanded to $\mathcal{S}'$. The scaled set $\mathcal{S}'$ consists of multiple copies of ranking vectors from $\mathcal{S}$, maintaining the original structure while increasing the dataset size, such that $n' = |\mathcal{S}'| \geq n$. A key advantage of \texttt{BnB} is that the number of nodes in the search tree remains unchanged across all such scaled instances, ensuring that the number of operations required to compute bounds remains constant. Consequently, \texttt{BnB} solves a scaled instance with $n'$ observation ranking vectors in approximately $2n' + 2d$ operations, demonstrating its computational advantage over the {\texttt{ES} method}, which has a time complexity of $\mathcal{O}(4\times 2\times n) = \mathcal{O}(8n)$. As $m$ increases, the number of nodes in the decision tree grows, making \texttt{BnB} less competitive than the {\texttt{ES} method} for exact clustering when $\epsilon = 0$. However, {increasing the controlling parameter $\epsilon$ helps enhance computational time of the \texttt{BnB} algorithm.} 

We conclude this section by presenting another computational advantage of \texttt{BnB}, establishing a worst-case bound on the \texttt{BnB} tree depth. Theorem \ref{thm:tree_size} leverages the structure of the input ranking vectors to show that when the observation ranking vectors are inherently clustered around distant centroids, the computational time of \texttt{BnB} can significantly decrease due to the reduced tree depth.

\begin{theorem}[Worst-Case Bound on the \texttt{BnB} Tree Depth] 
    \label{thm:tree_size} 
    Consider \texttt{KRC} with $n$ observation ranking vectors $\mathcal{S}$, $m$ options, and $2$ clusters. In a cluster reconstruction step of \texttt{KRCA}, let $\yy_1 = \begin{bmatrix} 1 & 2 & \dots & m \end{bmatrix}$ and $\yy_2 = \begin{bmatrix} m & m{-}1 & \dots & 1 \end{bmatrix}$. Assume all observation ranking vectors are within Euclidean distance $\delta \in \mathbb{Z}_+$ to one of the given centroids; that is, for all $\xx \in \mathcal{S}$, either $\|\xx - \yy_1\|^2 \le \delta^2$ or $\|\xx - \yy_2\|^2 \le \delta^2$. Assume \texttt{BnB} with $\epsilon = 0$ is applied to assign the observation ranking vectors to the centroids. Then, the depth of the \texttt{BnB} tree is at most $\mu(m, \delta)$, where $\mu(m, \delta)$ is the minimum of $m$ and
    \begin{align*}
    \min \left\{ \bar{m} \ \Bigg| \ 
    \underset{\substack{
    \xx\in\mathcal{X}_m,\\
    \sum_{j=1}^{\bar{m}} (x_j - j)^2 \le {\delta}^2
    }}{\max}
    \left(\sum_{j=1}^{m} (x_j - j)^2\right)
    \le
    \underset{\substack{
    \xx\in\mathcal{X}_m,\\
    \sum_{j=1}^{\bar{m}} (x_j - j)^2 \le {\delta}^2
    }}{\min} 
    \left(\sum_{j=1}^{\bar{m}} (x_j - (m - j + 1))^2\right)
    \right\}.
    \end{align*}
\end{theorem}

\color{black}

\begin{figure}[ht!]
  \centering
  \includegraphics[width=.4\textwidth, trim = 0cm 19cm 10cm 1cm]{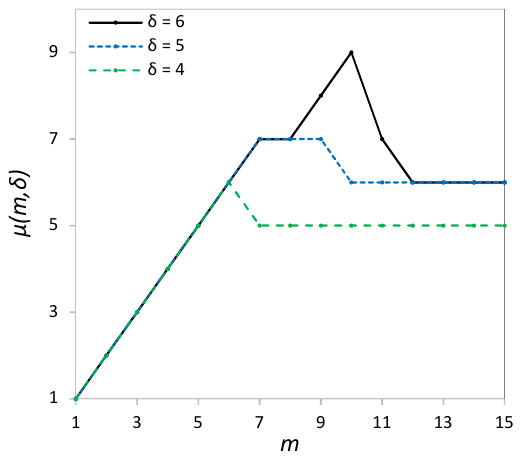}
  \caption{The worst-case tree-depth bound $\mu(m,\delta)$ versus the number of options $m$ for three values of $\delta \in \{4, 5, 6\}$.}
  \label{fig:BnBdepth}
\end{figure}



{Figure~\ref{fig:BnBdepth} plots the bound presented in Theorem~\ref{thm:tree_size}, demonstrating the following two insights. First, as $\delta$ decreases, the depth of the \texttt{BnB} tree decreases because the points of a cluster get closer to its center. Second, (beyond a certain $m$) as $m$ increases, the depth of the \texttt{BnB} tree decreases because the points of the other cluster move farther away.}
  
{In summary, we presented our efficient approximation algorithm, \texttt{KRCA}, designed to handle large-scale instances of \texttt{KRC}. A key contribution is the development of the \texttt{BnB} algorithm for the cluster reconstruction step, which is particularly effective when $m$ is small or the observation ranking vectors are inherently clustered. Additionally, \texttt{BnB} incorporates a hyperparameter, $\epsilon$, enabling users to balance computation time and solution quality. In the remainder, we numerically evaluate the computational performance of \texttt{KRCA} and highlight the practical advantages of \texttt{BnB}.}


\section{Numerical Experiments}
\label{sec:numerical_experiments}

In this section, we assess the efficacy of the \texttt{KRCA} algorithm on both synthetic {(Section \ref{subsec:synthetic_data}) and real datasets (Section \ref{subsec:real_data}).} To provide a basis for comparison, we use a two-step \texttt{KMC}-based procedure as the ``baseline'' solution. In the first step, the \texttt{kmeans} function in \texttt{MATLAB} is applied to cluster the given observation ranking vectors. The output consists of a {partition of the ranking vectors into clusters}, where each ranking vector is assigned to a cluster, and each cluster is represented by a centroid computed as the mean of the assigned ranking vectors. However, since the centroids obtained in this step may not necessarily be valid ranking vectors, the second step applies Theorem \ref{thm:optimal_centeriod} to determine {centroids in the form of ranking vectors}. This two-step procedure serves as the baseline for evaluating the improvements achieved by the \texttt{KRCA} algorithm. Additionally, as described in the initialization step of the \texttt{KRCA} algorithm {(Section \ref{sec:krca})}, this baseline solution is also used as the initial solution for \texttt{KRCA}.

The extensive numerical results reported in this section demonstrate the efficacy and novelty of the \texttt{KRCA} algorithm compared to the {baseline solution.} 
These findings highlight the significant contributions of the \texttt{KRCA} algorithm in addressing the limitations of the baseline approach. Moreover, the novel \texttt{BnB} algorithm, developed for the cluster reconstruction step of the \texttt{KRCA} algorithm {(Section~\ref{sec:krca}), significantly improves} the running time of \texttt{KRCA} compared to the {\texttt{ES} method}, depending on the problem size and parameter settings.

The numerical experiments were conducted on a Microsoft Windows 11 Enterprise operating system using \texttt{MATLAB} R2024b on a machine equipped with an Intel(R) Core(TM) i7-9700 CPU (3.00 GHz), featuring 8 cores and 8 logical processors, and 32 GB of RAM.

For the baseline solution, we set the maximum iteration limit of the \texttt{kmeans} function in \texttt{MATLAB} to $5,000$, meaning the algorithm iterates up to $5,000$ times to converge or stop when reaching the limit. For the \texttt{KRCA} algorithm, the stopping criterion is defined such that if the improvement in the objective value over two consecutive iterations is less than $10^{-6}$, the algorithm terminates. Additionally, unless otherwise stated, we set the hyperparameter $\epsilon = 10^{-6}$ in the \texttt{BnB} algorithm.

\subsection{Synthetic Data: Verification and Implementation}
\label{subsec:synthetic_data}

{We synthetically generate and analyze a variety of dataset instances to validate the theoretical results presented in Sections~\ref{sec:single-cluster_error_bounds} and~\ref{sec:krca}, and to evaluate the performance of the \texttt{KRCA} and \texttt{BnB} algorithms. Specifically, we conduct an in-depth numerical analysis of the \texttt{KRCA} algorithm (Section~\ref{subsubsec:uniformly_gen_rv}), compare the \texttt{BnB} and \texttt{ES} methods (Section~\ref{subsubsec:BnB_vs_ES}), and examine the effect of the hyperparameter~$\epsilon$ in the \texttt{BnB} algorithm (Section~\ref{subsubsec:effects_of_controlling_param}) on uniformly generated datasets. We also analyze the performance of the \texttt{BnB} algorithm on clustered ranking vectors (Section~\ref{subsubsec:clustered_rv}).}

\subsubsection{{Numerical Analysis of the \texttt{KRCA} Algorithm.}}
\label{subsubsec:uniformly_gen_rv}
To evaluate the performance of the \texttt{KRCA} algorithm, we simulate a large number of instances by uniformly generating ranking vector datasets. Here, ``uniformly'' refers to the condition where each permutation in the set $\mathcal{X}_m$ for a given $m$ is generated with equal probability, $1 / m!$. The number of ranking vectors in each dataset is set to $n \in \{100{,}000,\, 200{,}000,\, \dots,\, 1{,}000{,}000\}$, the number of options to $m \in \{4, \ldots, 8\}$, and the number of clusters to $k \in \{2, \ldots, 7\}$. Considering all combinations of $n$, $m$, and $k$, we generate $10 \times 5 \times 6 = 300$ ranking vector datasets. For each dataset, we first compute the baseline solution, which is then used as the initial solution to run the \texttt{KRCA} algorithm. {We repeat this process $10$ times for each dataset,} resulting in a total of $3{,}000$ solution pairs comprising the baseline and \texttt{KRCA} solutions. For each pair, the relative improvement in the objective value is calculated as follows: 
\begin{align}
    \label{eq:relative_improvement}
    \text{Relative Improvement} & = {\frac{v_{\mathtt{KRC}}(\bar{\mathcal{S}_1},\ldots,\bar{\mathcal{S}_k}) - v_{\mathtt{KRC}}(\mathcal{S}_1^{\circ},\ldots,\mathcal{S}_k^{\circ})}{v_{\mathtt{KRC}}(\mathcal{S}_1^{\circ},\ldots,\mathcal{S}_k^{\circ})}} \times 100\%,
\end{align}
where $\bar{\mathcal{S}_1}, \ldots, \bar{\mathcal{S}_k}$ denote the cluster sets produced by the baseline solution, and {$\mathcal{S}_1^{\circ}, \ldots, \mathcal{S}_k^{\circ}$} represent the cluster sets generated by the \texttt{KRCA} algorithm. In both cases, the centroids are computed according to Theorem \ref{thm:optimal_centeriod}. As explained earlier in this section, the baseline solution utilizes the \texttt{kmeans} function in \texttt{MATLAB}, which incorporates randomization in generating its initial solution. Consequently, different replications of the \texttt{kmeans} function may produce different baseline solutions.

{We use the \texttt{BnB} algorithm to reconstruct clusters within \texttt{KRCA} (Line 5 of Algorithm \ref{alg:generic_framework}) when $m \leq 5$, otherwise, the {\texttt{ES} method} is employed. This threshold was selected based on our extensive numerical comparison between \texttt{BnB} and \texttt{ES} on uniformly generated datasets (Section~\ref{subsubsec:BnB_vs_ES}).}

\begin{figure}[t]
  \centering
  \includegraphics[width=.7\textwidth, trim = 0cm 12.5cm 6cm 1cm]{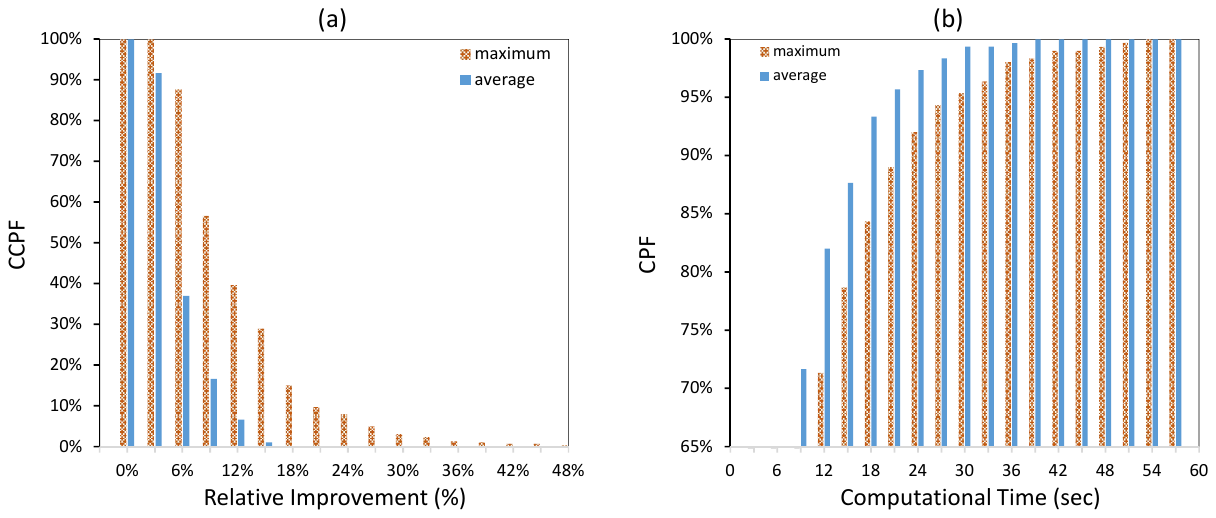}
  \caption{Figures (a) and (b) present summary statistics across all $300$ uniformly generated instances. {Figures (a) and (b) respectively display the \texttt{CCPF} histogram of relative improvements (Equation~\eqref{eq:relative_improvement}) and \texttt{CPF} histogram of computational times (in seconds) across replications.} 
  }
  \label{fig:uniform_vs_basline}
\end{figure}

We begin with an overall presentation of the results, followed by a stratified analysis based on the primary parameters $n$, $m$, and $k$. Figure \ref{fig:uniform_vs_basline} summarizes the percentage frequency across all $300$ uniformly generated datasets. Figure \ref{fig:uniform_vs_basline}(a) illustrates the complementary cumulative percentage frequency ($\mathtt{CCPF}$) histogram of the average and maximum relative improvements (Equation~\eqref{eq:relative_improvement}) across the $10$ replications for each dataset. It shows that in $88\%$ of the datasets, at least one of the $10$ replications achieved a relative improvement of at least $6\%$, and in $37\%$ of the datasets, the average relative improvement across the $10$ replications was at least $6\%$. Furthermore, in approximately $10\%$ of the datasets, at least one replication achieved a significant relative improvement of $20\%$ or more.
{These results demonstrate that \texttt{KRCA} achieves significant improvements over the baseline solution.} 
Figure \ref{fig:uniform_vs_basline}(b) depicts the cumulative percentage frequency ($\mathtt{CPF}$) histogram of the average and maximum computational times across the $10$ replications for each dataset. It shows that in $78\%$ of the datasets, all replications were solved within at most $15$ seconds, and in $82\%$ of the datasets, the average computational time was at most $12$ seconds. The close alignment of the two curves highlights the high stability of \texttt{KRCA}'s computational performance.

\begin{figure}[t]
  \centering
  \includegraphics[width=1\textwidth, trim = 0cm 13.5cm 0cm .5cm]{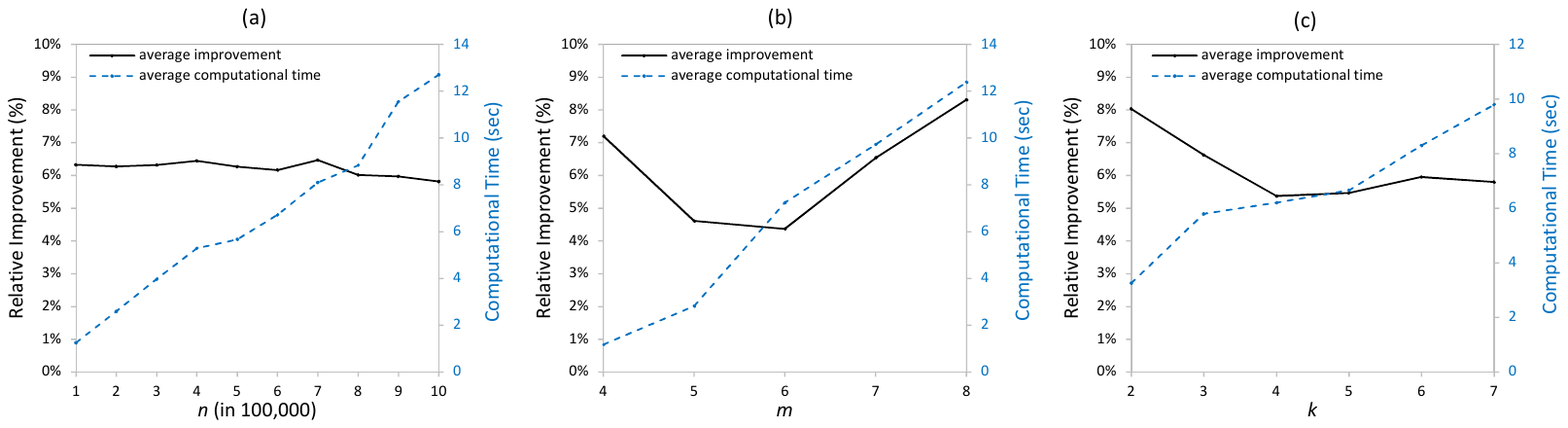}
  \caption{Figures (a), (b), and (c) show the effect of the number of observations $n$, the number of options $m$, and the number of clusters $k$, respectively, on the relative improvement (\%) and computational time (in seconds), using the \texttt{KRCA} algorithm. The relative improvement is plotted on the left vertical axis in black, and the computational time is plotted on the right vertical axis in blue.}
  \label{fig:uniform_nmk}
\end{figure}

Next, we analyze the effect of each parameter, namely, the number of observations $n$, the number of options $m$, and the number of clusters $k$, on relative improvement and computational time. Figure~\ref{fig:uniform_nmk} presents the average relative improvement (shown on the left vertical axis as a solid black line) and the average computational time in seconds (shown on the right vertical axis as a blue dashed line) across all generated instances for each fixed value of the parameter, using the \texttt{KRCA} algorithm. Figure \ref{fig:uniform_nmk}(a) shows that the average relative improvement remains largely stable as $n$ increases, while the computational time grows approximately linearly with $n$, demonstrating \texttt{KRCA}'s scalability for large datasets. In Figure \ref{fig:uniform_nmk}(b), the average relative improvement initially decreases as $m$ increases, likely due to reduced variability in lower-dimensional ranking vectors, but then increases again for larger values of $m$. Meanwhile, computational time increases moderately with $m$. Figure \ref{fig:uniform_nmk}(c) shows a slight decrease in relative improvement as $k$ increases, while computational time exhibits a {mild upward trend}. Overall, these findings highlight the robustness and computational efficiency of \texttt{KRCA} across a wide range of problem sizes.

In summary, Figures~\ref{fig:uniform_vs_basline} and~\ref{fig:uniform_nmk} demonstrate that the \texttt{KRCA} algorithm consistently outperforms the baseline solution in terms of relative improvements, while maintaining reasonable computational efficiency even for large-scale datasets. The relative improvement is largely independent of $n$, exhibits a non-monotonic trend with $m$, and decreases modestly with $k$. {Moreover, the computational time scales linearly with $n$ and increases moderately with $m$ and $k$.} 

\subsubsection{Comparison Between the \texttt{BnB} and \texttt{ES} Methods.}
\label{subsubsec:BnB_vs_ES}

\begin{figure}[t]
  \centering
  \includegraphics[width=1\textwidth, trim = 0cm 13.5cm 0cm .5cm]{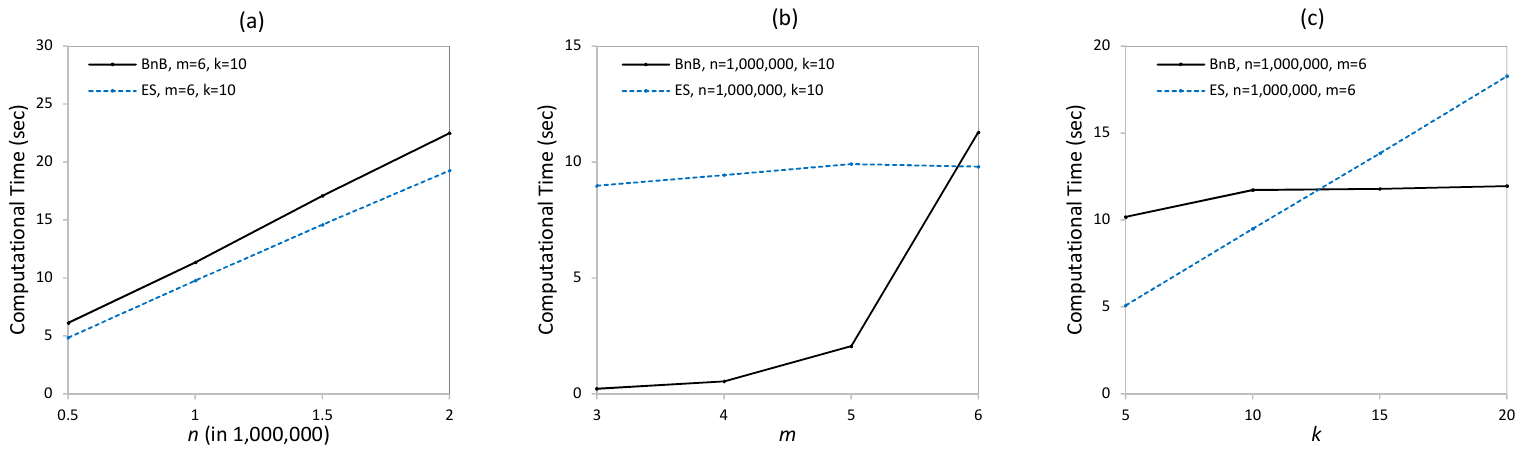}
  \caption{Figures (a), (b), and (c) show the effect of the number of observations $n$, the number of options $m$, and the number of clusters $k$, respectively, on the computational time (in seconds) using the \texttt{BnB} and {\texttt{ES} methods}. The computational time of \texttt{BnB} is plotted in black, and that of \texttt{ES} is plotted in blue.
  }
  \label{fig:BnB_BF_nmk}
\end{figure}

Figure \ref{fig:BnB_BF_nmk} presents a direct comparison between the \texttt{BnB} and {\texttt{ES} methods} for cluster reconstruction, evaluated on {uniformly generated instances. For each parameter setting, we uniformly generate $n$ ranking vectors and $k$ centroids, and then apply both the \texttt{BnB} and \texttt{ES} methods, with each experiment replicated 10 times.} Figure \ref{fig:BnB_BF_nmk}(a) shows the computational time {(average over the 10  replications)} as a function of the number of observations $n$ (in millions), with fixed values $m = 6$ and $k = 10$. The results indicate that increasing $n$ has little effect on the relative performance of the two methods. Although \texttt{BnB} is theoretically expected to benefit from larger $n$, this advantage is not strongly reflected in the observed computational times. Figure \ref{fig:BnB_BF_nmk}(b) varies the number of options $m$ while fixing $n = 1{,}000{,}000$ and $k = 10$. The results reveal a clear threshold: \texttt{BnB} outperforms \texttt{ES} for $m \leq 5$, while \texttt{ES} becomes more efficient for larger $m$. This cutoff has also been consistently observed across a range of parameter settings for uniformly generated datasets. Finally, Figure \ref{fig:BnB_BF_nmk}(c) shows the impact of the number of clusters $k$, with $n = 1{,}000{,}000$ and $m = 6$. In this case, increasing $k$ improves the relative performance of \texttt{BnB}, and there exists a threshold beyond which \texttt{BnB} becomes faster than \texttt{ES}. However, this crossover effect is less pronounced compared to the sensitivity with respect to $m$. Notably, for $m = 7$, we did not observe any value of $k$ for which \texttt{BnB} outperformed \texttt{ES}.

\subsubsection{Effects of the Controlling Parameter $\epsilon$ in the \texttt{BnB} Algorithm.}
\label{subsubsec:effects_of_controlling_param}

\begin{figure}[t]
  \centering
  \includegraphics[width=.7\textwidth, trim = 0cm 12.5cm 6cm 1cm]{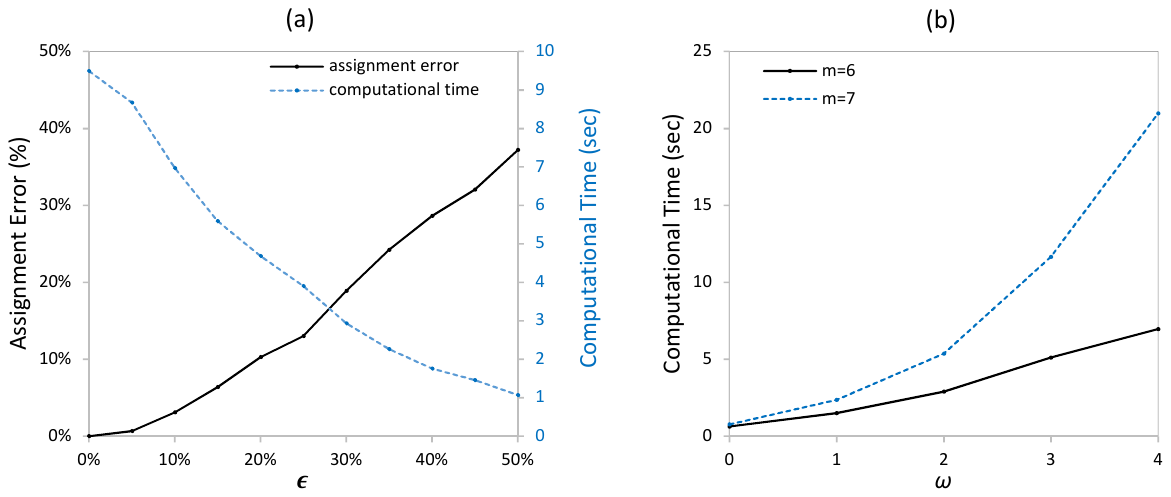}
  \caption{Figure (a) shows the assignment error (in black, left axis) and computational time (in blue, right axis) of the \texttt{BnB} algorithm for varying values of $\epsilon$, with $n = 1{,}000{,}000$, $m = 6$, and $k = 5$ under uniform data generation. Figure (b) displays the computational time (in seconds) as a function of the concentration parameter $\omega$, using the swapping iteration method with $n = 1{,}000{,}000$, $k = 5$, and $m \in \{6, 7\}$.
  }
  \label{fig:clustered_epsilon_fig}
\end{figure}

We examine the role of the hyperparameter $\epsilon$ in the \texttt{BnB} algorithm, focusing on its effect on computational efficiency and solution accuracy. As discussed in Section \ref{subsubsec:convergence_properties_bnb}, and particularly in Theorem \ref{thm:alg_b&b_error}, the \texttt{BnB} algorithm enables approximate assignment of ranking vectors to clusters through $\epsilon$. This allows a trade-off between faster execution and higher assignment precision. 
{Figure \ref{fig:clustered_epsilon_fig}(a) illustrates the behavior of \texttt{BnB} with respect to $\epsilon$. We generate $n = 1{,}000{,}000$ ranking vectors and $k = 5$ centroids uniformly, with $m = 6$ options.} The plot shows the average assignment error and corresponding computational time for various values of $\epsilon$. {The assignment error is measured as the percentage increase in the objective value when a given $\epsilon$ value is used, relative to the objective value obtained with $\epsilon = 0$.} As expected, increasing $\epsilon$ leads to a nearly linear growth in assignment error but a significant reduction in computational time. This empirical trend aligns closely with the theoretical bound in Theorem \ref{thm:alg_b&b_error}, confirming that larger $\epsilon$ values yield faster, though less precise, solutions.

\subsubsection{Clustered {Datasets}.}
\label{subsubsec:clustered_rv}
We extend our numerical analysis to scenarios where the input dataset is inherently clustered. Our aim is to investigate {the performance of the \texttt{BnB} method on clustered datasets, as discussed in Section \ref{subsubsec:convergence_properties_bnb} and Theorem \ref{thm:tree_size}.}

We begin by describing a procedure for randomly generating clustered ranking vectors. The process involves generating $k$ centroids and creating ranking vectors associated with each centroid, ensuring that different clusters remain sufficiently distinct. We first outline the method for generating a single cluster and then extend this approach to create multiple clusters under the specified assumptions. To generate a cluster, we randomly select a ranking vector $\yy \in \mathcal{X}_m$ as the centroid. Clustered ranking vectors are constructed by iteratively swapping consecutive entries of $\yy$. Specifically, at each step, a pair of consecutive integers $\ell, \ell+1 \in \{1, \ldots, m\}$ is randomly chosen, and their positions are swapped. This process is repeated $\omega$ times to produce a random ranking vector $\xx \in \mathcal{X}_m$. For example, consider the centroid $\yy = \begin{bmatrix} 1 & 2 & 3 & 4 \end{bmatrix} \in \mathcal{X}_4$ with $\omega = 2$. In the first iteration, if $2$ and $3$ are selected, swapping them results in $\begin{bmatrix} 1 & 3 & 2 & 4 \end{bmatrix}$. In the second iteration, if $1$ and $2$ are selected, the final ranking vector becomes $\xx = \begin{bmatrix} 2 & 3 & 1 & 4 \end{bmatrix}$. The squared Euclidean distance between $\xx$ and $\yy$ is given by $\|\xx - \yy\|^2 = 6$. Proposition \ref{pro:bound_on_omega_iterations} establishes an upper bound on the squared Euclidean distance between the centroid and any ranking vector generated after $\omega$ iterations. This bound ensures that the generated vectors stay within a controlled distance from their centroids, maintaining cluster cohesion. Additionally, it ensures that the randomly generated vectors are approximately uniformly distributed within a controlled distance from the given centroid \citep{Diaconis1981GeneratingAR}.

\begin{proposition}[Upper Bound on Clustered Vector Distance]
    \label{pro:bound_on_omega_iterations}
    Let $\yy$ be a given centroid in $\mathcal{X}_m$, and let the iterative procedure described above be applied $\omega$ times to $\yy$ to generate a random ranking vector $\xx$. For $\omega < m$, it almost surely holds that,
    \begin{align*}
        \| \xx - \yy \|^2 & \leq 2 \omega^2.
    \end{align*}
\end{proposition}
To create multiple clusters for a given parameter $\omega$, {following the insight from Proposition \ref{pro:bound_on_omega_iterations}, we generate $k$ centroids} with distances greater than $2\omega^2$. Next, one of the $k$ centroids is randomly selected, and the iterative random swapping procedure is applied to generate a ranking vector belonging to the corresponding cluster. This process is repeated $n$ times to generate all clustered ranking vectors.  This approach helps maintain distinct cluster identities while allowing for partial overlap, reflecting realistic scenarios where input data naturally forms clusters that are not entirely disjoint. Additionally, the {expected sizes of the clusters are equal.}

{Figure~\ref{fig:clustered_epsilon_fig}(b) displays the computational time of the \texttt{BnB} method on clustered datasets, with $n = 1{,}000{,}000$, $k = 5$, and $m \in \{6, 7\}$. The parameter $\omega$ controls} the concentration of clusters: smaller $\omega$ values result in fewer swaps and hence more compact, highly concentrated clusters, while larger values lead to greater dispersion. The results show that as $\omega$ decreases, the computational time of \texttt{BnB} drops sharply. In particular, for highly concentrated datasets (e.g., $\omega \leq 1$), \texttt{BnB} completes the assignment step in just a few seconds. These findings align with the {tree-depth} characterization in Theorem \ref{thm:tree_size} and highlight the algorithm's ability to exploit data regularity for improved efficiency.

Thus far, we have presented numerical analyses on generated synthetic datasets, including instances with uniformly generated ranking vectors and clustered ranking vectors. In the remainder of this section, we shift our focus to experiments on a real dataset.

\subsection{Real Data: MovieLens Dataset}
\label{subsec:real_data}

In this section, we use the MovieLens dataset \citep{harpermovielens} to evaluate the performance of the \texttt{KRCA} algorithm in {personalization} based on users' genre-preference rankings. We begin by introducing the MovieLens dataset and outlining the procedure for deriving ranking vector datasets from users' assigned movie ratings. This process generates 10 benchmark ranking vector datasets for evaluating the performance of \texttt{KRCA}. We then apply \texttt{KRCA} to these benchmark datasets and analyze the resulting relative improvements and computational times. Overall, our experiments with the MovieLens dataset demonstrate the efficacy of \texttt{KRCA} and highlight its practical applicability in real-world scenarios.

\subsubsection{Overview of the MovieLens Dataset.}
\label{subsubsec:overview_of_movieLens_dataset}
We used the dataset file titled ``ml-25m.zip'' from \citet{grouplensmovielens} and combined the relevant files to create a comprehensive dataset of over $20$ million observation records, containing ratings from $138{,}493$ users for $26{,}744$ movies. Each record represents a rating assigned by a user to a movie and includes the following fields: user ID, assigned rating, movie ID, movie name, and genres. The ratings range from $0.5$ to $5$, with an overall average of $3.52$ and a standard deviation of $1.05$. Of the total observation records, $83.10\%$ correspond to movies labeled with more than one genre, $15.55\%$ correspond to movies labeled with a single genre, and a small remaining portion lack genre information. In this study, we focus on observation records corresponding to movies labeled with a single genre, resulting in a streamlined dataset containing $3{,}109{,}550$ records. This subset includes movies from the four most frequent genres: Drama, Comedy, Documentary, and Thriller.

\begin{figure}[t]
  \centering
  \includegraphics[width=1\textwidth, trim = 0cm 14cm 1cm .5cm]{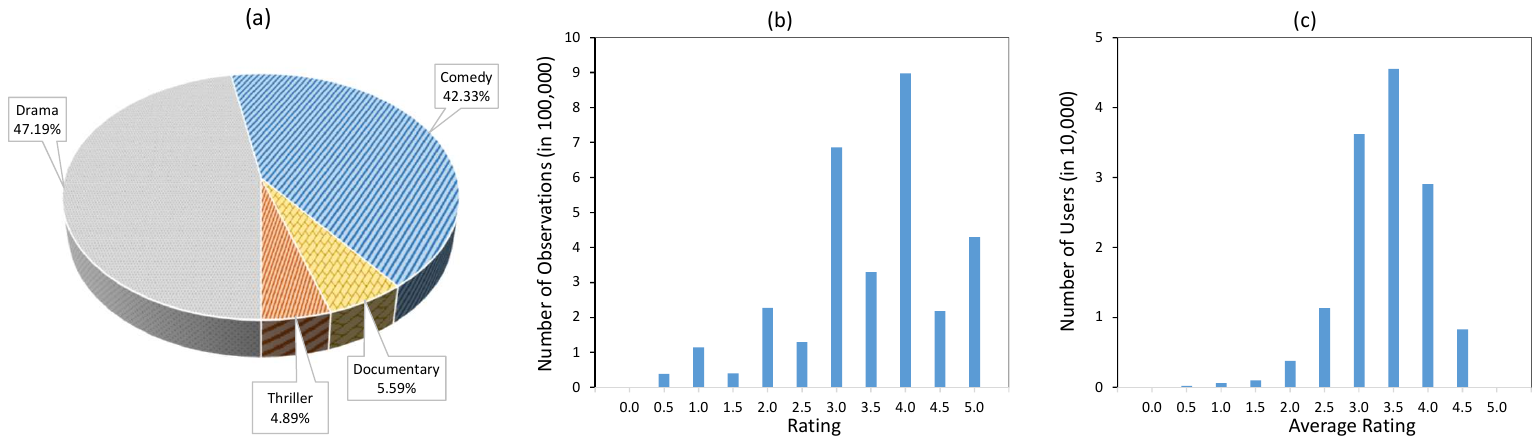}
  \caption{Figures (a), (b), and (c) show the pie chart of the distribution of the four genres, the histogram of ratings across all observation records, and the histogram of average ratings, where each average is calculated per user based on all their ratings in the streamlined dataset used in this study.}
  \label{fig:summary_charts}
\end{figure}

Figure \ref{fig:summary_charts} illustrates the distribution of key features in the streamlined dataset. Figure \ref{fig:summary_charts}(a) shows a pie chart of the percentage of observation records for each genre, highlighting that Drama and Comedy dominate, collectively accounting for $89.52\%$ of the total records. Figure \ref{fig:summary_charts}(b) presents a histogram of all ratings, revealing a {multimodal} distribution with prominent modes at $3.0$ and $4.0$, and additional minor modes around $2.0$ and $5.0$. Figure \ref{fig:summary_charts}(c) displays a histogram of average ratings per user, showing a unimodal distribution with a peak at $3.5$ and a slight {leftward skew.} 

\subsubsection{Generating Benchmark Ranking Vector Datasets.}
\label{subsubsec:generating_benchmark_rvs}
We introduce the $\lambda$-filtering procedure, given in Algorithm \ref{alg:lambda_procedure}, to construct benchmark datasets for the \texttt{KRC} problem using {the streamlined dataset described in Section~\ref{subsubsec:overview_of_movieLens_dataset}.} The $\lambda$-filtering procedure uses the hyperparameter $\lambda$ to filter the dataset, retaining users who have rated at least $\lambda$ movies in every genre $g \in \{1, 2, \dots, m\}$, where $m = 4$ in this study. Higher $\lambda$ values ensure that only users with more comprehensive ratings across genres are retained, improving data quality. For each retained user $i$, the average rating $r_i^g$ is computed for each genre $g$, and the deviation $\Delta r_i^g = r_i^g - \bar{r}^g$ is calculated, where $\bar{r}^g$ is the overall average rating for genre $g$ across all users in the streamlined dataset. {We interpret this deviation $\Delta r_i^{g}$ as} the additional utility user $i$ derives from movies of genre $g$ relative to $\bar{r}^g$. Sorting the $\Delta r_i^g$ values in descending order produces a ranking vector $\xx_i$ for user $i$, where the $g^{\text{th}}$ coordinate of $\xx_i$ indicates the rank of $\Delta r_i^g$. In the case of ties (equal $\Delta r_i^g$ values for multiple genres), an arbitrary order may be used. In summary, Algorithm \ref{alg:lambda_procedure} converts the streamlined dataset into a set of ranking vectors determined by the filtering parameter $\lambda$.
\begin{algorithm}[h!]
    \OneAndAHalfSpacedXI\small\footnotesize
    \begin{algorithmic}[1]
        \caption{\small The $\lambda$-Filtering Procedure
        \label{alg:lambda_procedure}}
        \Begin
        \State \textsc{Inputs:} Streamlined dataset $\mathcal{D}$, total number of genres $m$, filtering parameter $\lambda$
        \State Initialize ranking vectors set $\mathcal{R}_\lambda = \emptyset$
        
        \For{(each user $i$ in $\mathcal{D}$)}
            \If{(user $i$ has rated at least $\lambda$ movies in every genre $g \in \{1, 2, \dots, m\}$)}
                \For{(each genre $g \in \{1, 2, \dots, m\}$)}
                    \State Compute $r_i^{g} =$ average rating user $i$ assigns to movies of genre $g$
                    \State Compute $\Delta r_i^{g} = r_i^{g} - \bar{r}^{g}$, where $\bar{r}^{g}$ is the overall average rating for genre $g$ across all users in $\mathcal{D}$
                \EndFor
                \State Generate the ranking vector $\xx_i$ by ranking $\Delta r_i^{g}$ values in descending order for $g \in \{1, 2, \dots, m\}$
                \State Add the generated ranking vector $\xx_i$ to $\mathcal{R}_\lambda$
            \EndIf
        \EndFor

        \State \textsc{Outputs:} The set of ranking vectors $\mathcal{R}_\lambda$
        \End
    \end{algorithmic}
\end{algorithm}

\begin{figure}[t]
  \centering
  \includegraphics[width=.7\textwidth, trim = 0cm 19.5cm 0cm .5cm]{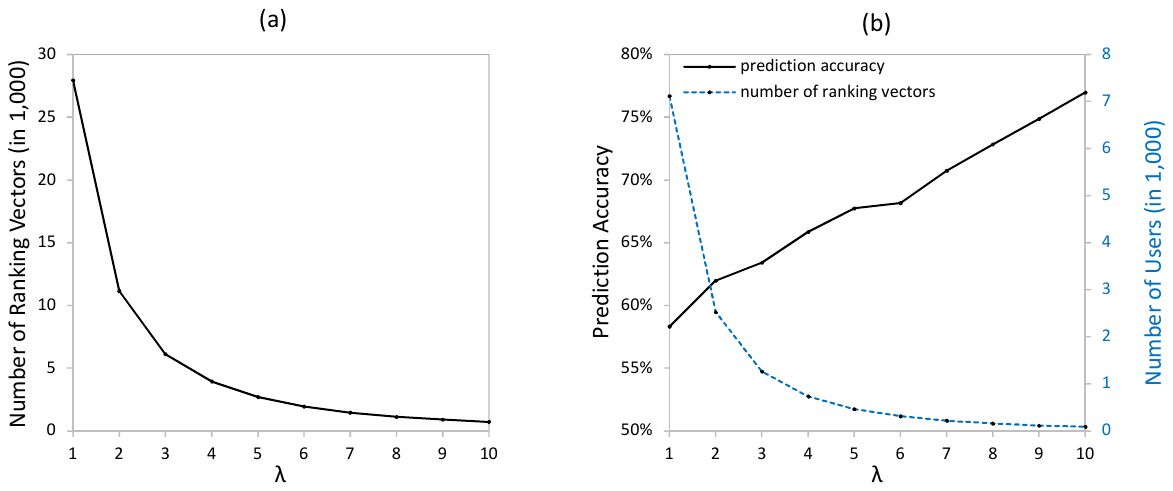}
  \caption{Figure (a) displays the number of generated ranking vectors (in $1{,}000$) across a range of $\lambda$ values from $1$ to $10$. Figure (b) illustrates the prediction accuracy for the test set (\%, left vertical axis in black) and the number of remaining users after the double-filtering process (in $1{,}000$, right vertical axis in blue) for the same range of $\lambda$ values.}  
  \label{fig:benchmark_datasets}
\end{figure}

We apply Algorithm \ref{alg:lambda_procedure} to generate $10$ different ranking vector datasets by setting $\lambda \in \{1, 2, \dots, 10\}$. Figure \ref{fig:benchmark_datasets}(a) illustrates the number of generated ranking vectors in each dataset for different values of $\lambda$. As $\lambda$ increases, the number of users, and consequently the number of generated ranking vectors in the filtered dataset, decreases. These $10$ datasets collectively serve as a benchmark real-world dataset to evaluate the practical performance of the \texttt{KRCA} algorithm.

\subsubsection{Rationale for the $\lambda$-Filtering Procedure.}
\label{subsubsec:rationale_lambda_filtering_procedure}
We provide the rationale for the $\lambda$-filtering procedure by analyzing its impact on accurately predicting users' genre-preference rankings. To do this, we randomly split the streamlined dataset, which contains $3{,}109{,}550$ observations, into training and testing sets by assigning each observation to one of the two sets with equal probability. The $\lambda$-filtering procedure is then applied separately to each set, resulting in ranking vectors for the users remaining after filtering. To ensure consistency between the training and testing sets, we retain only the users common to both sets, discarding those present in only one. This double-filtering process ensures that both sets contain identical users, although their ranking vectors may differ due to variations in the observation records within each set. Next, we solve the \texttt{KRC} problem using the training set, clustering the users into $k$ clusters and determining their optimal centroids. For each user in the testing set, we calculate the squared Euclidean distances to all $k$ centroids obtained from the training set. A prediction is considered ``correct'' if the shortest distance corresponds to the same cluster the user was assigned to in the training set; otherwise, it is considered ``incorrect''. The percentage of correctly predicted users in the testing set serves as the measure of prediction accuracy, validating the effectiveness of the $\lambda$-filtering procedure.

We set $k=2$ and vary $\lambda$ from $1$ to $10$, performing $10$ replications for each value and averaging the prediction results across replications. Figure \ref{fig:benchmark_datasets}(b) shows that as $\lambda$ increases, the number of double-filtered users decreases, while prediction accuracy improves significantly within the given range. For example, when $\lambda=1$, the average number of remaining double-filtered users is $7{,}113$, with a prediction accuracy of $58.33\%$. Increasing $\lambda$ to $5$ reduces the number of users to $467$, with accuracy improving to $67.74\%$. Doubling $\lambda$ to $10$ further reduces the number of users to $89$, while accuracy increases to $76.96\%$. These results demonstrate that $\lambda$ acts as a threshold balancing the trade-off between the number of retained users and prediction accuracy. Larger $\lambda$ values result in higher accuracy but reduce the number of retained users due to stricter double-filtering. These results demonstrate the effectiveness of the $\lambda$-filtering procedure in improving prediction accuracy.

\subsubsection{Performance of \texttt{KRCA} on the Benchmark Datasets.}
\label{subsubsec:performance_krc_on_benchmark_datasets}
We use the 10 generated benchmark datasets, where the number of ranking vectors $n$ ranges from $719$ to $27{,}937$, with $m = 4$ genres, to evaluate the performance of the \texttt{KRCA} algorithm. We set the number of clusters $k \in \{3, 4, \dots, 7\}$ and, for each $k$, compute the baseline solutions, followed by applying {the \texttt{KRCA} algorithm, with 10 replications.} This results in a total of $10 (\text{datasets}) \times 5 (\text{cluster values}) \times 10 (\text{replications}) = 500$ pairs of solutions. 

\begin{figure}[t]
  \centering
  \includegraphics[width=.7\textwidth, trim = 0cm 12.5cm 6cm 1cm]{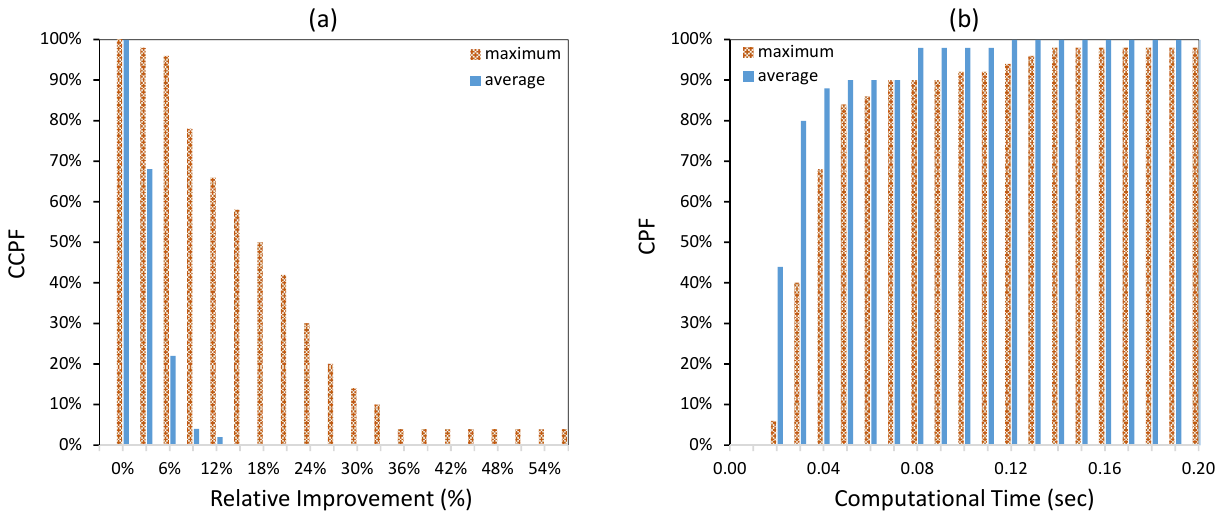}
  \caption{Improvements achieved by \texttt{KRCA} and its computational time on the benchmark instances.}\label{fig:benchmark_histograms}
\end{figure}

Analogous to Figure \ref{fig:uniform_vs_basline}, Figure \ref{fig:benchmark_histograms} summarizes the percentage frequency across all 500 {pairs of solutions. Figure \ref{fig:benchmark_histograms}(a) presents the \texttt{CCPF} histogram of the average and maximum relative improvements.} It shows that in 96\% of the datasets, at least one of the 10 replications achieved a relative improvement of at least 5\%, and in 48\% of the datasets, the average relative improvement across the 10 replications was at least 5\%. Additionally, in approximately 25\% of the datasets, at least one replication achieved a significant relative improvement of 25\%. 
{These results underscore} the importance of developing the novel \texttt{KRCA} algorithm.

Figure \ref{fig:benchmark_histograms}(b) presents the \texttt{CPF} histogram of the average and maximum computational times across the 10 replications for each dataset. It shows that in $92\%$ of the datasets, all replications were solved within at most $0.1$ seconds. The close alignment of the two curves in Figure \ref{fig:benchmark_histograms}(b) highlights the stability of \texttt{KRCA}'s computational performance. 

{In summary, our extensive} numerical experiments on both synthetic and real-world datasets demonstrate the effectiveness of the \texttt{KRCA} algorithm in {achieving significant improvements in solution quality compared to the baseline method.} Moreover, \texttt{KRCA} consistently achieves fast computational times, underscoring its scalability for large datasets. These results demonstrate the practical significance of \texttt{KRCA} in addressing real-world ranking vectors clustering problems while ensuring robust computational performance. The alignment of the numerical findings in this section with the theoretical results established in Sections \ref{sec:single-cluster_error_bounds} and \ref{sec:krca} further validates the reliability and relevance of the proposed algorithm, making it a valuable tool for solving \texttt{KRC} problems in diverse applications.


\section{Conclusion and Future Work}
\label{sec:conclusion_future_work}

{This paper presents a robust framework for clustering ranking vectors, referred to as \texttt{KRC}, with broad applicability ranging from personalization to large-scale group decision making. \texttt{KRC} enables both personalized recommendations, offers, and promotions, and the summarization of collective preferences, thereby supporting more effective decisions.}

{We present a comprehensive theoretical study of \texttt{KRC}, prove its \texttt{NP}-hardness, and derive a closed-form solution for the optimal centroid of each cluster. To address the computational challenges, we introduce the \texttt{KRCA} algorithm, which iteratively refines initial solutions derived from \texttt{KMC}, along with a novel \texttt{BnB} algorithm for efficient cluster reconstruction.}

{Our extensive numerical experiments on synthetic and real-world datasets demonstrate that \texttt{KRCA} significantly improves solution quality while maintaining fast computational times. These results highlight the scalability and efficiency of \texttt{KRCA} for large datasets, validating its practical utility in real-world applications. Additionally, the alignment of empirical findings with theoretical results reinforces the reliability of our proposed methodology.}

In the future, several directions could be explored to extend this work. First, incorporating dynamic ranking preferences, where user preferences evolve over time, could enhance the applicability of \texttt{KRCA} to real-time systems. Second, extending \texttt{KRCA} to handle weighted ranking vectors, where some preferences carry more significance than others, could broaden its use in multi-criteria decision making problems. Finally, investigating distributed or parallel implementations of \texttt{KRCA} would further improve its scalability for handling massive datasets in real-time applications.



\newpage
\bibliographystyle{informs2014}
\bibliography{Ref}

\clearpage\newpage

\pagenumbering{arabic} 
\setcounter{page}{1}

\begin{center}{\large

    \

   \textbf{Ranking Vectors Clustering: Theory and Applications}} 
   
   \vspace{3mm} 
   
   {\large
   
   (Online Appendix)}

   \vspace{3mm} 

\end{center}

\begin{APPENDICES}
\section{Technical Lemmas and Proofs}
\label{sec:appendix}

\subsection{Proof of Lemma \ref{lem:characterization_ranking_vectors}}
\textit{Proof. }We need to show that for any ranking vector $\xx \in \mathcal{X}_m$, it satisfies $\xx \in \mathsf{BHS}\left(\mathbf{0},\frac{1}{6}m(m+1)(2m+1)\right)$ and $\xx \in \mathsf{HP}\left(\mathbf{1},\frac{1}{2}m(m+1)\right)$. This can be verified through straightforward algebraic calculations using the following two fundamental properties of ranking vectors:
\begin{align}
    \label{eq:<x,1>} \langle \xx, \mathbf{1} \rangle & = \sum_{i=1}^{m} i = \frac{1}{2}m(m+1),
\end{align}
and
\begin{align}
    \label{eq:norm(x,2)} \| \xx \|^2 & = \sum_{i=1}^{m} i^2 = \frac{1}{6}m(m+1)(2m+1).
\end{align}
Equation \eqref{eq:<x,1>} gives the inner product of the vectors $\xx$ and $\mathbf{1}$, ensuring that $\xx$ lies in $\mathsf{HP}\left(\mathbf{1},\frac{1}{2}m(m+1)\right)$. Similarly, Equation \eqref{eq:norm(x,2)} represents the square of the Euclidean distance of $\xx$ from the origin, confirming that $\xx$ lies within $\mathsf{BHS}\left(\mathbf{0},\frac{1}{6}m(m+1)(2m+1)\right)$. Therefore, both conditions are satisfied.
\qed

\subsection{Proof of Theorem \ref{thm:NPhardness}}
To prove Theorem \ref{thm:NPhardness}, we first introduce the following definitions, lemma, and theorem.

\begin{definition}[Hypercube Clustering Problem \citep{baldi2012boolean}]
    \label{def:hcp}
    Consider a set ${\mathcal{S}} = \{\zz_1, \ldots, \zz_n\}$, where each $\zz_i \in \{0, 1\}^\eta$ is an $\eta$-dimensional binary vector for $i = 1, \ldots, n$. The ``Hypercube Clustering Problem'' (\texttt{HCP}) is defined as finding clusters ${\mathcal{S}}_1, \dots, {\mathcal{S}}_k$ and corresponding centroids $\ww_1, \dots, \ww_k \in \{0, 1\}^\eta$ such that the total Hamming distance,
    \begin{align*}
        \sum_{\ell=1}^k \sum_{\zz \in \mathcal{S}_\ell} f_{\mathtt{H}}(\zz, \ww_\ell),
    \end{align*}
    is minimized. Here, $f_{\mathtt{H}}(\zz, \ww_\ell)$ denotes the \textit{Hamming distance} between $\zz$ and $\ww_\ell$, defined as the number of positions where the entries of $\zz$ and $\ww_\ell$ differ.
\end{definition}

\begin{theorem}[\texttt{NP}-Hardness of \texttt{HCP} \citep{baldi2012boolean}]
    \label{thm:np_hard_hcp}
    The \texttt{HCP} is \texttt{NP}-hard.
\end{theorem}

\begin{definition}[Alternating Pair Transformation]
    \label{def:alter_pair_transform}
    Let $\zz = \begin{bmatrix}
        z_1 & \cdots & z_\eta
    \end{bmatrix} \in \{ 0,1 \}^\eta$ be a binary vector. The alternating pair transformation maps $\zz$ to an $m$-dimensional vector $\xx = \begin{bmatrix}
    x_{1} & \cdots & x_{m}
    \end{bmatrix}$, where $m = 2\eta$, as follows:
    \begin{align*}
        x_{j} & = 
        \begin{cases}
            j + z_{(j+1)/2}, & \text{for } j = 1, 3, \ldots, m-1, \\
            j - z_{j/2}, & \text{for } j = 2, 4, \ldots, m.
        \end{cases}
    \end{align*}
\end{definition}

\begin{lemma}[Alternating Pair Ranking Vector]
    \label{lem:alter_pair_rv}
    The alternating pair transformation maps any binary vector $\zz \in \{ 0,1 \}^\eta$ to a vector $\xx \in \mathcal{X}_m$, where $m = 2\eta$, called an ``alternating pair ranking vector''.
\end{lemma}

\noindent\textit{Proof. }From Definition \ref{def:alter_pair_transform}, since $z_j \in \{ 0,1 \}$ for all $j$, each pair of successive coordinates $(x_j, x_{j+1})$ is either $(j, j+1)$ or $(j+1, j)$ for $j = 1, 3, \ldots, m-1$. Consequently, $\xx$ satisfies the properties of a ranking vector, and therefore, $\xx \in \mathcal{X}_m$.
\qed

\begin{definition}[Inverse Alternating Pair Mapping]
    \label{def:inv_alter_pair_struc}
    Let $\xx = \begin{bmatrix}
        x_1 & \cdots & x_m
    \end{bmatrix} \in \mathcal{X}_m$ be an alternating pair ranking vector. The unique $\eta$-dimensional binary vector $\zz$, where $\eta = m / 2$, can be derived using the inverse alternating pair mapping as follows:
    \begin{align*}
        z_j & = 
        \begin{cases}
            0 & \text{if } (x_{2j-1}, x_{2j}) = (2j-1, 2j), \\
            1 & \text{if } (x_{2j-1}, x_{2j}) = (2j, 2j-1),
        \end{cases}
    \end{align*}
    for $j = 1, \ldots, \eta$.
\end{definition}

\noindent\textit{Proof of Theorem \ref{thm:NPhardness}. }To prove the \texttt{NP}-hardness of the \texttt{KRC} problem, we demonstrate that any arbitrary instance of the \texttt{HCP} can be transformed into an instance of the \texttt{KRC} problem in linear time with respect to the problem parameters. Furthermore, we show that an optimal clustering of the transformed \texttt{KRC} instance can be used to derive an optimal clustering for the original \texttt{HCP} instance, also in linear time. Hence, by Theorem \ref{thm:np_hard_hcp}, we conclude that the \texttt{KRC} problem is \texttt{NP}-hard.

\noindent Consider the instance $\mathtt{HCP}^\circ$, consisting of $n$ $\eta$-dimensional binary vectors $\zz_1^\circ, \ldots, \zz_n^\circ$ and a clustering number $k$. The corresponding instance $\mathtt{KRC}^\circ$ is constructed by transforming these binary vectors into $n$ $m$-dimensional alternating pair ranking vectors $\xx_1^\circ, \ldots, \xx_n^\circ$, where $m = 2\eta$, while preserving the clustering number $k$. We first show that an optimal clustering for the $\mathtt{KRC}^\circ$ problem consists of $k$ clusters, with all $k$ optimal centroids being alternating pair ranking vectors. Furthermore, applying the inverse alternating pair mapping to the optimal centroids, while maintaining the same cluster assignments (i.e., all alternating pair ranking vectors $\xx_j^\circ$ within a cluster map to their corresponding binary vectors $\zz_j^\circ$, which also belong to the same cluster), results in an optimal clustering for the instance $\mathtt{HCP}^\circ$. Since the alternating pair transformation and its inverse can both be performed in linear time with respect to $n$, $m$, and $k$, this completes the proof.

\noindent Let $\mathcal{S}_1^*, \ldots, \mathcal{S}_k^*$ be an optimal clustering for the $\mathtt{KRC}^\circ$ instance. The average of the vectors assigned to cluster $\mathcal{S}_\ell^*$, denoted as $\bar{\xx}_\ell = \begin{bmatrix}
    \bar{x}_{\ell 1} & \cdots & \bar{x}_{\ell m}
\end{bmatrix}$, is given by
\begin{align*}
    \bar{\xx}_\ell & = \frac{1}{| \mathcal{S}_\ell^* |} \sum_{\xx^\circ \in \mathcal{S}_\ell^*} \xx^\circ, \quad \text{for } \ell = 1, \ldots, k.
\end{align*}
Due to the structure of alternating pair ranking vectors, this average has the property that, for successive pairs of coordinates, we have
\begin{align*}
    (\bar{x}_{\ell j}, \bar{x}_{\ell j+1}) & = \left( \frac{c_{\ell j} j + d_{\ell j}(j+1)}{c_{\ell j} + d_{\ell j}}, \frac{c_{\ell j} (j+1) + d_{\ell j} j}{c_{\ell j} + d_{\ell j}} \right),
    \quad \text{for } j = 1, 3, \ldots, m-1,
\end{align*}
where $c_{\ell j}$ and $d_{\ell j}$ represent the counts of pairs $(j, j+1)$ and $(j+1, j)$, respectively, in the $j^{\text{th}}$ and $(j+1)^{\text{st}}$ coordinates of all vectors in cluster $\mathcal{S}_\ell^*$. Furthermore, all successive pairs $(\bar{x}_{\ell j}, \bar{x}_{\ell j+1})$ are ordered in ascending order relative to one another. Specifically, for any two odd indices $1\leq j_1 < j_2 < m$, both values in the pair $(\bar{x}_{\ell j_1}, \bar{x}_{\ell j_1+1})$ are smaller than those in the pair $(\bar{x}_{\ell j_2}, \bar{x}_{\ell j_2+1})$. Thus, by applying Theorem \ref{thm:optimal_centeriod}, the optimal centroid $\yy_\ell^* = \begin{bmatrix}
    y_{\ell 1}^* & \cdots & y_{\ell m}^*
\end{bmatrix}$ for cluster $\mathcal{S}_\ell^*$ is an alternating pair ranking vector, with the following values for $j = 1, 3, \ldots, m-1$:
\begin{itemize}
    \item If $c_{\ell j} < d_{\ell j}$, then $\bar{x}_{\ell j} > \bar{x}_{\ell j+1}$, implying $y_{\ell j}^* = j+1$ and $y_{\ell j+1}^* = j$.
    \item If $c_{\ell j} \geq d_{\ell j}$, then $\bar{x}_{\ell j} \leq \bar{x}_{\ell j+1}$, implying $y_{\ell j}^* = j$ and $y_{\ell j+1}^* = j+1$.
\end{itemize}

\noindent Finally, we conclude the proof by demonstrating that applying the inverse alternating pair mapping to the optimal centroids $\yy_1^*, \ldots, \yy_m^*$, while preserving the same clusters, results in an optimal clustering for the instance $\mathtt{HCP}^\circ$. According to Definitions \ref{def:alter_pair_transform} and \ref{def:inv_alter_pair_struc}, there exists a one-to-one correspondence between the set of all $\eta$-dimensional binary vectors and the set of all $m$-dimensional alternating pair ranking vectors, where $m = 2\eta$. This correspondence ensures that any vector from one set uniquely determines a vector in the other set. Additionally, for any binary vectors $\zz, \ww \in \{0,1\}^\eta$ and their corresponding alternating pair ranking vectors $\xx, \yy \in \mathcal{X}_m$, the following cases hold for $j = 1, \ldots, \eta$:
\begin{description}
    \item[Case 1:] If $z_j = 0$ and $w_j = 0$, then $(x_{2j-1}, x_{2j}) = (2j-1, 2j)$ and $(y_{2j-1}, y_{2j}) = (2j-1, 2j)$, so
    \[
    | z_j - w_j | = \frac{1}{2} \left( (x_{2j-1} - y_{2j-1})^2 + (x_{2j} - y_{2j})^2 \right) = 0.
    \]
    \item[Case 2:] If $z_j = 0$ and $w_j = 1$, then $(x_{2j-1}, x_{2j}) = (2j-1, 2j)$ and $(y_{2j-1}, y_{2j}) = (2j, 2j-1)$, so
    \[
    | z_j - w_j | = \frac{1}{2} \left( (x_{2j-1} - y_{2j-1})^2 + (x_{2j} - y_{2j})^2 \right) = 1.
    \]
    \item[Case 3:] If $z_j = 1$ and $w_j = 0$, then $(x_{2j-1}, x_{2j}) = (2j, 2j-1)$ and $(y_{2j-1}, y_{2j}) = (2j-1, 2j)$, so
    \[
    | z_j - w_j | = \frac{1}{2} \left( (x_{2j-1} - y_{2j-1})^2 + (x_{2j} - y_{2j})^2 \right) = 1.
    \]
    \item[Case 4:] If $z_j = 1$ and $w_j = 1$, then $(x_{2j-1}, x_{2j}) = (2j, 2j-1)$ and $(y_{2j-1}, y_{2j}) = (2j, 2j-1)$, so
    \[
    | z_j - w_j | = \frac{1}{2} \left( (x_{2j-1} - y_{2j-1})^2 + (x_{2j} - y_{2j})^2 \right) = 0.
    \]
\end{description}
Thus, it follows that:
\begin{align*}
    f_{\mathtt{H}}(\zz, \ww) & = \frac{1}{2} \| \xx - \yy \|^2.
\end{align*}
Therefore, for a given set of $k$ clusters with centroids $\ww_1^\circ, \ldots, \ww_k^\circ$ in the $\mathtt{HCP}^\circ$ instance and the corresponding alternating pair ranking vectors $\yy_1^\circ, \ldots, \yy_k^\circ$ in the $\mathtt{KRC}^\circ$ instance, we have:
\begin{align*}
    \sum_{\ell=1}^k \sum_{i \in \mathcal{I}_\ell} f_{\mathtt{H}}(\zz_i^\circ, \ww_\ell^\circ) & = \frac{1}{2} \sum_{\ell=1}^k \sum_{i \in \mathcal{I}_\ell} \| \xx_i^\circ - \yy_\ell^\circ \|^2,
\end{align*}
where $\mathcal{I}_\ell$ denotes the set of indices of vectors assigned to the $\ell^{\text{th}}$ cluster. This demonstrates that the optimal objective value of $\mathtt{HCP}^\circ$ is half the optimal objective value of $\mathtt{KRC}^\circ$. Considering the one-to-one correspondence between binary vectors and alternating pair ranking vectors, along with this equality in objective values, we conclude that $\mathtt{HCP}^\circ$ and $\mathtt{KRC}^\circ$ solve equivalent optimization problems. Consequently, an optimal clustering for one instance can be directly mapped to an optimal clustering for the other. This completes the proof.
\qed

\subsection{Proof of Theorem \ref{thm:optimal_centeriod}}
\textit{Proof. }We prove this theorem by applying a dynamic programming approach and utilizing the induction method. The aim is to solve the following optimization problem:
\begin{align*}
    \min_{\yy\in\mathcal{X}_m} v_{\mathtt{KRC}}(\mathcal{S}, \yy) & = \min_{\yy\in\mathcal{X}_m} \sum_{\xx\in\mathcal{S}} \| \xx - \yy \|^2 \\
    & = \min_{\yy\in\mathcal{X}_m} \sum_{i=1}^{n} \| \xx_i - \yy \|^2 \\
    & = \min_{\yy\in\mathcal{X}_m} \sum_{i=1}^{n}\sum_{j=1}^{m} (x_{ij} - y_j)^2.
\end{align*}
To address this, let us define 
\begin{align*}
    u_k(\mathcal{R}_k) & \eq \min\ \sum_{j=1}^{k}\sum_{i=1}^{n} (x_{ij} - y_j)^2 \\
    \text{subject to } &\quad y_j \in \mathcal{R}_k, \\
    \text{and } &\quad y_j \neq y_{j'}\quad \text{for } j\neq j'\in\{1,\ldots,k\}, 
\end{align*}
where $\mathcal{R}_k \subseteq \{1,\ldots, m\}$ consists of $k$ distinct integer numbers. Clearly,
\begin{align*}
    u_m(\mathcal{R}_m) = \min_{\yy\in\mathcal{X}_m} v_{\mathtt{KRC}}(\mathcal{S}, \yy).    
\end{align*}
By utilizing the Bellman optimality equation \citep{ross1992dynamic}, we have 
\begin{align}
    \label{eq:bellman_g} u_k(\mathcal{R}_k) & = \min_{y_k\in \mathcal{R}_k} \left\{ \sum_{i=1}^{n} (x_{ik} - y_k)^2 + u_{k-1}(\mathcal{R}_k\backslash \{y_k\}) \right\},
\end{align}
with the termination condition
\begin{align}
    \label{eq:bellman_i} u_1(\mathcal{R}_1) & = \sum_{i=1}^{n} (x_{i1} - r_1)^2,
\end{align}
where $\mathcal{R}_1 = \{r_1\}$ is a singleton.

\noindent We first show that the statement in Theorem \ref{thm:optimal_centeriod} holds for $k = 2$. Without loss of generality, assume that $\bar{x}_{.1} \leq \bar{x}_{.2}$ and $\mathcal{R}_2 = \{ r_1, r_2 \}$, where $1 \leq r_1 < r_2 \leq m$. By applying \eqref{eq:bellman_g} and \eqref{eq:bellman_i}, we have
\begin{align*}
    u_2(\mathcal{R}_2) & = \min_{y_2 \in \mathcal{R}_2} \left\{ \sum_{i=1}^{n} (x_{i2} - y_2)^2 + u_1(\mathcal{R}_2 \setminus \{y_2\}) \right\} \\
    & = \min \left\{ \sum_{i=1}^{n} (x_{i2} - r_2)^2 + \sum_{i=1}^{n} (x_{i1} - r_1)^2,\ \sum_{i=1}^{n} (x_{i2} - r_1)^2 + \sum_{i=1}^{n} (x_{i1} - r_2)^2 \right\}.
\end{align*}
To compare these two terms, consider the difference
\begin{align*}
    & \left(\sum_{i=1}^{n} (x_{i2} - r_2)^2 + \sum_{i=1}^{n} (x_{i1} - r_1)^2 \right) 
    - \left(\sum_{i=1}^{n} (x_{i2} - r_1)^2 + \sum_{i=1}^{n} (x_{i1} - r_2)^2 \right) \\
    & \quad\quad\quad\quad = \left(\sum_{i=1}^{n} (x_{i2} - r_2)^2 - \sum_{i=1}^{n} (x_{i2} - r_1)^2 \right) 
    - \left(\sum_{i=1}^{n} (x_{i1} - r_2)^2 - \sum_{i=1}^{n} (x_{i1} - r_1)^2 \right) \\
    & \quad\quad\quad\quad = \sum_{i=1}^{n} \left[ (r_1 - r_2)(2x_{i2} - r_1 - r_2) \right] 
    - \sum_{i=1}^{n} \left[ (r_1 - r_2)(2x_{i1} - r_1 - r_2) \right] \\
    & \quad\quad\quad\quad = \sum_{i=1}^{n} \left[ 2(r_1 - r_2)x_{i2} - (r_1^2 - r_2^2) \right] 
    - \sum_{i=1}^{n} \left[ 2(r_1 - r_2)x_{i1} - (r_1^2 - r_2^2) \right] \\
    & \quad\quad\quad\quad = 2n(r_1 - r_2)(\bar{x}_{.2} - \bar{x}_{.1}).
\end{align*}
Since $\bar{x}_{.1} \leq \bar{x}_{.2}$ and $r_1 < r_2$, it follows that
\begin{align*}
    2n(r_1 - r_2)(\bar{x}_{.2} - \bar{x}_{.1}) \leq 0.
\end{align*}
Thus, we conclude
\begin{align*}
    \sum_{i=1}^{n} (x_{i2} - r_2)^2 + \sum_{i=1}^{n} (x_{i1} - r_1)^2 
    & \leq \sum_{i=1}^{n} (x_{i2} - r_1)^2 + \sum_{i=1}^{n} (x_{i1} - r_2)^2,
\end{align*}
which implies that $y_1^* = r_1$ and $y_2^* = r_2$ form an optimal solution for $u_2(\mathcal{R}_2)$.

\noindent Now, let us assume that the statement of Theorem \ref{thm:optimal_centeriod} holds for all $j < k < m$, and we show that it holds for $j = k$ as well. Without loss of generality, assume $\bar{x}_{.1} \leq \bar{x}_{.2} \leq \cdots \leq \bar{x}_{.k}$ and $\mathcal{R}_{k} = \{ r_1, \ldots, r_{k} \}$ for $1 \leq r_1 < r_2 < \cdots < r_{k} \leq m$. By utilizing the induction hypothesis, we have
\begin{align*}
    u_{k}(\mathcal{R}_{k}) & = \min_{y_{k}\in \mathcal{R}_{k}} \left\{ \sum_{i=1}^{n} (x_{i k} - y_k)^2 + u_{k-1}(\mathcal{R}_k\backslash \{y_k\}) \right\} \\
    & = \min_{r_{\ell}\in \mathcal{R}_{k}} \left\{ \sum_{i=1}^{n} (x_{i k} - r_\ell)^2 + \sum_{j=1}^{\ell-1} \sum_{i=1}^{n} (x_{ij} - r_{j})^2 + \sum_{j=\ell}^{k-1} \sum_{i=1}^{n} (x_{ij} - r_{j+1})^2 \right\}.
\end{align*}
We aim to show that $y_j = r_j$ for $j = 1, \ldots, k$ is an optimal solution for $u_{k}(\mathcal{R}_{k})$. To do so, we construct the following difference:
\begin{align*}
    & \left( \sum_{j=1}^{k} \sum_{i=1}^{n} (x_{ij} - r_{j})^2 \right) - \left( \sum_{i=1}^{n} (x_{i k} - r_\ell)^2 + \sum_{j=1}^{\ell-1} \sum_{i=1}^{n} (x_{ij} - r_{j})^2 + \sum_{j=\ell}^{k-1} \sum_{i=1}^{n} (x_{ij} - r_{j+1})^2 \right) \\
    & \quad\quad = \left( \sum_{j=1}^{\ell-1} \sum_{i=1}^{n} (x_{ij} - r_{j})^2 + \sum_{j=\ell}^{k-1} \sum_{i=1}^{n} (x_{ij} - r_{j})^2 + \sum_{i=1}^{n} (x_{ik} - r_{k})^2 \right) \\ 
    & \quad\quad\quad\quad - \left( \sum_{i=1}^{n} (x_{i k} - r_\ell)^2 + \sum_{j=1}^{\ell-1} \sum_{i=1}^{n} (x_{ij} - r_{j})^2 + \sum_{j=\ell}^{k-1} \sum_{i=1}^{n} (x_{ij} - r_{j+1})^2 \right) \\
    & \quad\quad = \left( \sum_{i=1}^{n} (x_{i k} - r_k)^2 - \sum_{i=1}^{n} (x_{i k} - r_\ell)^2 \right) + \left( \sum_{j=\ell}^{k-1} \sum_{i=1}^{n} (x_{ij} - r_{j})^2 -  \sum_{j=\ell}^{k-1} \sum_{i=1}^{n} (x_{ij} - r_{j+1})^2 \right) \\
    & \quad\quad = \sum_{i=1}^{n} \left[ (r_{\ell} - r_k)(2x_{i k} - r_\ell - r_k)\right] + \sum_{j=\ell}^{k-1}  \sum_{i=1}^{n} \left[(r_{j+1} - r_{j})(2x_{ij} - r_j - r_{j+1})\right] \\
    & \quad\quad = \sum_{i=1}^{n} \left[ 2(r_{\ell} - r_k)x_{i k} - (r_\ell^2 - r_k^2)\right] + \sum_{j=\ell}^{k-1}  \sum_{i=1}^{n} \left[2(r_{j+1} - r_{j})x_{ij} - (r_j^2 - r_{j+1}^2)\right] \\
    & \quad\quad = 2n(r_\ell - r_k)\bar{x}_{.k} - n(r_\ell^2 - r_k^2) + \sum_{j=\ell}^{k - 1}2n(r_{j+1} - r_j)\bar{x}_{.j} - \sum_{j=\ell}^{k - 1} n(r_{j+1}^2 - r_j^2) \\
    & \quad\quad = \left( 2n(r_\ell - r_k)\bar{x}_{.k} + \sum_{j=\ell}^{k - 1}2n(r_{j+1} - r_j)\bar{x}_{.j} \right) - \left( n(r_\ell^2 - r_k^2) + \sum_{j=\ell}^{k - 1} n(r_{j+1}^2 - r_j^2) \right) \\
    & \quad\quad = 2n(r_\ell - r_k)\bar{x}_{.k} + \sum_{j=\ell}^{k - 1}2n(r_{j+1} - r_j)\bar{x}_{.j}.
\end{align*}
Since $\bar{x}_{.1} \leq \bar{x}_{.2} \leq \cdots \leq \bar{x}_{.k}$ and $r_1 < r_2 < \cdots < r_k$, it follows:
\begin{align*}
    2n(r_\ell - r_k)\bar{x}_{.k} + \sum_{j=\ell}^{k - 1}2n(r_{j+1} - r_j)\bar{x}_{.j} & \leq 2n(r_\ell - r_k)\bar{x}_{.k} + \sum_{j=\ell}^{k - 1}2n(r_{j+1} - r_j)\bar{x}_{.k} \\
    & = 2n \bar{x}_{.k} \left( (r_\ell - r_k) + \sum_{j=\ell}^{k - 1}(r_{j+1} - r_j) \right) \\
    & = 0.
\end{align*}
Hence, we conclude that
\begin{align*}
    \sum_{j=1}^{k} \sum_{i=1}^{n} (x_{ij} - r_{j})^2 & \leq \sum_{i=1}^{n} (x_{i k} - r_\ell)^2 + \sum_{j=1}^{\ell-1} \sum_{i=1}^{n} (x_{ij} - r_{j})^2 + \sum_{j=\ell}^{k-1} \sum_{i=1}^{n} (x_{ij} - r_{j+1})^2,
\end{align*}
which implies that $y_i^* = r_i$ for $i=1, \ldots,k$ form an optimal solution for $u_k(\mathcal{R}_k)$. This completes the proof.
\qed

\subsection{Proof of Proposition \ref{pro:upper_bnd}}
In order to prove Proposition \ref{pro:upper_bnd}, we first introduce the following Lemmas. 

\begin{lemma}[Objective Value Relationship Between \texttt{KRC} and \texttt{KMC}]
    \label{lem:v_krc_vs._v_kmc}
    Let $\mathcal{S} = \{ \xx_1, \ldots, \xx_n \} \in \mathcal{X}_m$ represent $n$ ranking vectors partitioned into $k$ clusters $\mathcal{S}_1, \ldots, \mathcal{S}_k$. Then,
    \begin{align*}
        v_{\mathtt{KRC}}(\mathcal{S}_1, \ldots, \mathcal{S}_k) & = v_{\mathtt{KMC}}(\mathcal{S}_1, \ldots, \mathcal{S}_k) + \sum_{\ell=1}^{k} \sum_{\xx \in \mathcal{S}_{\ell}} \| \yy_{\mathtt{KRC},\ell}^* - \yy_{\mathtt{KMC},\ell}^* \|^2,
    \end{align*}
    where $\yy_{\mathtt{KRC},\ell}^*$ and $\yy_{\mathtt{KMC},\ell}^*$ are the optimal centroids of $\mathcal{S}_\ell$ for the \texttt{KRC} and \texttt{KMC} problems, respectively, for $\ell = 1, \ldots, k$.
\end{lemma}
\noindent\textit{Proof. }Applying Equation \eqref{eq:objFn_krc}, we have,
\begin{align*}
    v_{\mathtt{KRC}}(\mathcal{S}_1, \ldots, \mathcal{S}_k) & = \sum_{\ell=1}^{k}\sum_{\xx \in \mathcal{S}_{\ell}} \| \xx - \yy_{\mathtt{KRC},\ell}^* \|^2 \\
    & = \sum_{\ell=1}^{k}\sum_{\xx \in \mathcal{S}_{\ell}} \| \xx - \yy_{\mathtt{KMC},\ell}^* + \yy_{\mathtt{KMC},\ell}^* - \yy_{\mathtt{KRC},\ell}^* \|^2 \\
    & = \sum_{\ell=1}^{k}\sum_{\xx \in \mathcal{S}_{\ell}} \left( \| \xx - \yy_{\mathtt{KMC},\ell}^* \|^2 + \| \yy_{\mathtt{KMC},\ell}^* - \yy_{\mathtt{KRC},\ell}^* \|^2 + 2 \langle \xx - \yy_{\mathtt{KMC},\ell}^*, \yy_{\mathtt{KMC},\ell}^* - \yy_{\mathtt{KRC},\ell}^* \rangle \right).
\end{align*}
Recalling that vectors are assumed to be row vectors. From \eqref{eq:optimal_center_kmc}, we know,
\begin{align*}
    \yy_{\mathtt{KMC},\ell}^* & = \frac{1}{| \mathcal{S}_\ell |} \sum_{\xx \in \mathcal{S}_\ell} \xx,
\end{align*}
where $|\mathcal{S}_\ell|$ is the size of cluster $\mathcal{S}_\ell$. Using this result, it follows from standard linear algebra \citep{golub2013matrix} that,
\begin{align*}
    \sum_{\xx \in \mathcal{S}_{\ell}} \langle \xx - \yy_{\mathtt{KMC},\ell}^*, \yy_{\mathtt{KMC},\ell}^* - \yy_{\mathtt{KRC},\ell}^* \rangle & = 0,
\end{align*}
for $ \ell = 1, \ldots, k $. Thus, the cross-term vanishes, and from \eqref{eq:objFn_kmc}, we obtain,
\begin{align*}
    v_{\mathtt{KRC}}(\mathcal{S}_1, \ldots, \mathcal{S}_k) & = v_{\mathtt{KMC}}(\mathcal{S}_1, \ldots, \mathcal{S}_k) + \sum_{\ell=1}^{k}\sum_{\xx \in \mathcal{S}_{\ell}} \| \yy_{\mathtt{KMC},\ell}^* - \yy_{\mathtt{KRC},\ell}^* \|^2.
\end{align*}
This completes the proof.
\qed

\begin{lemma}[Maximum Dot Product with a Ranking Vector]
    \label{lem:max_dot_product_ranking_vector}
    Let $\xx \in \mathbb{R}^m_{+}$ be a positive vector. Define the ranking vector $\yy \in \mathcal{X}_m$ such that its $i^{\text{th}}$ component represents the rank of the $i^{\text{th}}$ coordinate of $\xx$ when the elements of $\xx$ are sorted in ascending order. For example, if a component of $\yy$ equals one, it corresponds to the smallest element of $\xx$. In the case of ties (equal coordinates in $\xx$), the ranking is determined based on their order of appearance in $\xx$. Then,
    \begin{align*}
        \max_{\cc \in \mathcal{X}_m} \langle \xx, \cc \rangle & = \langle \xx, \yy \rangle.
    \end{align*}
\end{lemma}
\noindent\textit{Proof. }Let $\xx = \begin{bmatrix}
    x_1 & \cdots & x_m 
\end{bmatrix}$ and $\yy = \begin{bmatrix}
    y_1 & \cdots & y_m 
\end{bmatrix}$. From the definition, we know that for any two indices $1 \leq i < j \leq m$, if $x_i \leq x_j$, then $y_i < y_j$. Now, consider a ranking vector $\cc  = \begin{bmatrix}
    c_1 & \cdots & c_m 
\end{bmatrix} \in \mathcal{X}_m$ such that there exist indices $1 \leq i < j \leq m$ where $x_i \leq x_j$ but $c_i > c_j$. We can show that,
\begin{align*}
    (c_j x_i + c_i x_j) - (c_i x_i + c_j x_j) & = (c_i - c_j)(x_j - x_i) \geq 0.
\end{align*}
This result implies that by exchanging $c_i$ and $c_j$ in the ranking vector $\cc$, we can improve the inner product with $\xx$. Hence, for any given ranking vector $\cc$, by iteratively applying this exchange procedure, we can improve the inner product until we reach $\yy$, where no further exchanges are possible. At this point, $\yy$ represents the ranking vector that maximizes the inner product. This completes the proof.
\qed

\noindent\textit{Proof of Proposition \ref{pro:upper_bnd}. }From Lemma \ref{lem:v_krc_vs._v_kmc}, it is readily seen that if we can show
\begin{align*}
    \sum_{\ell=1}^{k}\sum_{\xx \in \mathcal{S}_{\ell}} \| \yy_{\mathtt{KMC},\ell}^* - \yy_{\mathtt{KRC},\ell}^* \|^2 & \leq v_{\mathtt{KMC}}(\mathcal{S}_1, \ldots, \mathcal{S}_k),
\end{align*}
then the proof is complete. We have,
\begin{align*}
    \sum_{\ell=1}^{k}\sum_{\xx \in \mathcal{S}_{\ell}} \| \yy_{\mathtt{KMC},\ell}^* - \yy_{\mathtt{KRC},\ell}^* \|^2 & = \sum_{\ell=1}^{k}\sum_{\xx \in \mathcal{S}_{\ell}} \left( \| \yy_{\mathtt{KMC},\ell}^* \|^2 + \| \yy_{\mathtt{KRC},\ell}^* \|^2 - 2 \langle \yy_{\mathtt{KMC},\ell}^*, \yy_{\mathtt{KRC},\ell}^* \rangle \right).
\end{align*}
Applying Theorem \ref{thm:optimal_centeriod} together with Lemma \ref{lem:max_dot_product_ranking_vector}, we know that the inner product $\langle \yy_{\mathtt{KRC},\ell}^*, \yy_{\mathtt{KMC},\ell}^* \rangle$ is greater than or equal to $\langle \xx, \yy_{\mathtt{KMC},\ell}^* \rangle$ for any other ranking vector $\xx \in \mathcal{X}_m$. On the other hand, since $\yy_{\mathtt{KRC},\ell}^* \in \mathcal{X}_m$, we have for any ranking vector $\xx \in \mathcal{X}_m$,
\begin{align*}
    \| \yy_{\mathtt{KRC},\ell}^* \|^2 & = \| \xx \|^2 = \sum_{i=1}^{m} i^2.
\end{align*}
Thus, we can conclude that,
\begin{align*}
    \sum_{\ell=1}^{k}\sum_{\xx \in \mathcal{S}_{\ell}} \| \yy_{\mathtt{KMC},\ell}^* - \yy_{\mathtt{KRC},\ell}^* \|^2 & \leq \sum_{\ell=1}^{k}\sum_{\xx \in \mathcal{S}_{\ell}} \left( \| \yy_{\mathtt{KMC},\ell}^* \|^2 + \| \xx \|^2 - 2 \langle \yy_{\mathtt{KMC},\ell}^*, \xx \rangle \right) \\
    & = \sum_{\ell=1}^{k}\sum_{\xx \in \mathcal{S}_{\ell}} \| \xx - \yy_{\mathtt{KMC},\ell}^* \|^2 \\
    & = v_{\mathtt{KMC}}(\mathcal{S}_1, \ldots, \mathcal{S}_k).
\end{align*}
This completes the proof.
\qed

\subsection{Proof of Corollary \ref{cor:upper_bnd}}
\textit{Proof. }Let $\mathcal{S}_1^*, \ldots, \mathcal{S}_k^*$ represent an optimal clustering for the given \texttt{KRC} problem, and let $\mathcal{S}_1^\circ, \ldots, \mathcal{S}_k^\circ$ represent an optimal clustering for the corresponding \texttt{KMC} problem. Using the optimality of the \texttt{KRC} solution and applying Proposition \ref{pro:upper_bnd}, we obtain:
\begin{align*}
    v_{\mathtt{KRC}}^* & = v_{\mathtt{KRC}}(\mathcal{S}_1^*, \ldots, \mathcal{S}_k^*) \\
    & \leq v_{\mathtt{KRC}}(\mathcal{S}_1^\circ, \ldots, \mathcal{S}_k^\circ) \\
    & \leq 2 v_{\mathtt{KMC}}(\mathcal{S}_1^\circ, \ldots, \mathcal{S}_k^\circ) \\
    & = 2 v_{\mathtt{KMC}}^*.
\end{align*}
The lower bound $v_{\mathtt{KRC}}^* \geq v_{\mathtt{KMC}}^*$ is straightforward since \texttt{KRC} is a more constrained version of the \texttt{KMC} problem, with its feasible region for choosing the centroids being a subset of the feasible region of \texttt{KMC}. This completes the proof.
\qed

\subsection{Proof of Proposition \ref{pro:eps-optimal_krc}}
\textit{Proof. }Applying Proposition \ref{pro:upper_bnd} and the $\varepsilon$-optimality of the given solution for \texttt{KMC}, we have:
\begin{align*}
    v_{\mathtt{KRC}}(\bar{\mathcal{S}}_1, \ldots, \bar{\mathcal{S}}_k) & \leq 2 v_{\mathtt{KMC}}(\bar{\mathcal{S}}_1, \ldots, \bar{\mathcal{S}}_k) \\
    & \leq 2(v_{\mathtt{KMC}}^*+\varepsilon v_{\mathtt{KMC}}^*) \\
    & \leq 2(1+\varepsilon)v_{\mathtt{KRC}}^* \\
    & = v_{\mathtt{KRC}}^* + (1+2\varepsilon) v_{\mathtt{KRC}}^*.
\end{align*}
This completes the proof.
\qed

\subsection{Proof of Theorem \ref{thm:alg_b&b_error}}

Let $\phi'$ be the pruned node in the \texttt{BnB} tree with $\mathcal{S}_{\phi'} \neq \varnothing$ and 
$\mathcal{C}_{\phi'} = \{\yy_k\}$. Thus, \texttt{BnB} assigns every $\xx \in \mathcal{S}_{\phi'}$ to $\yy_k$ and then prunes $\phi'$. Along the unique route from the root node $\phi_0$ to $\phi'$, the remaining centers 
$\yy_{k-1}, \dots, \yy_1$ are dismissed one at a time (lines 18--22 of Algorithm \ref{alg:b&b}). Denote by $\phi_\ell$ the node at which $\yy_\ell$ is dismissed for $\ell = 1, \dots, k-1$. 

\noindent For every $\ell \in \{1, \dots, k-1\}$, we show that
\begin{align}
    \mathtt{UB}_{\phi_{k-1}, \yy_k} & \leq \mathtt{LB}_{\phi_{k - \ell}, \yy_{k - \ell}} + \ell \epsilon.
    \label{eq:UB-LB}
\end{align}
When $\ell = 1$, center $\yy_{k-1}$ is dismissed in $\phi_{k-1}$, so the pruning test gives
\begin{align*}
    \mathtt{UB}_{\phi_{k-1}, \yy_k} & \leq \mathtt{LB}_{\phi_{k-1}, \yy_{k-1}} + \epsilon,
\end{align*}
establishing \eqref{eq:UB-LB} for $\ell = 1$. 

\noindent Assume \eqref{eq:UB-LB} holds for all integers $2 \leq \ell \leq k-1$. Since $\yy_{k - \ell}$ is dismissed in $\phi_{k - \ell}$,
\begin{align*}
    \min_{s \in \{1, \dots, \ell\}} 
    \mathtt{UB}_{\phi_{k - \ell}, \yy_{k - \ell + s}} & \leq \mathtt{LB}_{\phi_{k - \ell}, \yy_{k - \ell}} + \epsilon.
\end{align*}
If the minimizer is $s = \ell$, the desired bound follows immediately. Otherwise, if $s = \tilde{\ell} < \ell$, the monotonicity of upper bounds along ancestor nodes together with the induction hypothesis for $\ell - \tilde{\ell}$ yields \eqref{eq:UB-LB}. Thus, \eqref{eq:UB-LB} holds for every $\ell$.

\noindent Fix any $\xx \in \mathcal{S}_{\phi'}$. Applying \eqref{eq:UB-LB} with $\ell = k - 1$ and noting that $\yy_{k - \ell}$ ranges over all centers in $\mathcal{C}$ gives
\begin{align}
    \| \xx - \yy_k \|^{2} & \leq \| \xx - \yy \|^{2} + (k-1)\epsilon
    \quad \text{for all } \yy \in \mathcal{C}.
    \label{eq:key-x}
\end{align}

\noindent For each $\xx \in \mathcal{S}$, let $\yy_{\xx}$ be the center selected by \texttt{BnB} and $\yy_{\xx}^*$ the center in the optimal partition $(\mathcal{S}_1^*, \dots, \mathcal{S}_k^*)$. If $\xx \in \mathcal{S}_{\phi'}$, apply \eqref{eq:key-x} with $\yy = \yy_{\xx}^*$, otherwise, $\yy_{\xx} = \yy_{\xx}^*$ and the inequality holds trivially. Hence,
\begin{align*}
    \| \xx - \yy_{\xx} \|^{2} & \leq \| \xx - \yy_{\xx}^* \|^{2} + (k-1)\epsilon 
    \quad \text{for all } \xx \in \mathcal{S}.
\end{align*}
Summing over all $n$ observations yields
\begin{align*}
    v_{\mathtt{KRC}}(\bar{\mathcal{S}}_1, \dots, \bar{\mathcal{S}}_k, \yy_1, \dots, \yy_k) 
    - v_{\mathtt{KRC}}(\mathcal{S}_1^*, \dots, \mathcal{S}_k^*, \yy_1, \dots, \yy_k) 
    & \leq n (k-1) \epsilon,
\end{align*}
completing the proof.
\qed

\subsection{Proof of Corollary \ref{cor:convergence_bnb}}
\textit{Proof. }The proof follows directly by setting $\epsilon = 0$ in Theorem \ref{thm:alg_b&b_error}.
\qed

\subsection{Proof of Theorem \ref{thm:tree_size}}
\textit{Proof. }Fix any level $\bar m\in\{1,\dots,m\}$ of the \texttt{BnB} decision tree and let $\phi$ be an arbitrary node on that level. By construction, $\phi$ contains  
\begin{align*}
    \bar{m} \text{ cuts of the form }x_{\hat j} & =j\quad \text{for } \hat{j} \in \{1,\dots,\bar m\},
\end{align*}
so all ranking vectors in $\mathcal{S}_\phi$ share the first $\bar m$ coordinates $f_\phi=(x_1,\dots,x_{\bar m})$. Because \texttt{BnB} only creates non-empty nodes, $\mathcal{S}_\phi\neq\varnothing$. Moreover, every $\xx\in \mathcal{S}_\phi$ is within squared Euclidean distance $\delta^2$ of at least one of the two apposite centroids
\begin{align*}
    \yy_1 & = \begin{bmatrix}
        1 & 2 & \cdots & m
    \end{bmatrix}, \\
    \yy_2 & = \begin{bmatrix}
        m & m-1 & \cdots & 1
    \end{bmatrix}.
\end{align*}
We establish the result in four steps, by analyzing the structure of any node at depth $\bar{m}$ and showing under what condition it is guaranteed to be pruned.

\paragraph{Step 1: Bounds with respect to $\yy_1$.}
Assume first that $\mathcal{S}_\phi$ contains at least one vector whose closest centroid is $\yy_1$. We have $\mathtt{LB}_{\phi,\yy_1} \leq \delta^2$. To decide whether $\phi$ can be pruned, we compare the largest upper bound attainable for $\yy_1$ with the smallest lower bound attainable for $\yy_2$ among all nodes whose first $\bar{m}$ coordinates satisfy the above constraint. Define
\begin{align*}
    \mathtt{UUB}_{\phi,\yy_1} & := \max_{\substack{\xx\in\mathcal {X}_m\\ \sum_{j=1}^{\bar m}(x_j-j)^2\le\delta^2}} \sum_{j=1}^{m}(x_j-j)^2, \\
    \mathtt{LLB}_{\phi,\yy_2} & := \min_{\substack{\xx\in\mathcal{X}_m\\ \sum_{j=1}^{\bar m}(x_j-j)^2\le\delta^2}} \sum_{j=1}^{\bar m}\bigl(x_j-(m-j+1)\bigr)^2,
\end{align*}
where, $\xx = \begin{bmatrix}
        x_1 & \cdots & x_m
    \end{bmatrix}$.

\paragraph{Step 2: Bounds with respect to $\yy_2$ and symmetry.}
Assume instead that $\mathcal{S}_\phi$ contains a point within $\delta$ of $\yy_2$. Symmetric definitions give
\begin{align*}
    \mathtt{UUB}_{\phi,\yy_2} & := \max_{\substack{\xx\in\mathcal{X}_m\\ \sum_{j=1}^{\bar m}(x_j-(m-j+1))^2\le\delta^2}} \sum_{j=1}^{m}\bigl(x_j-(m-j+1)\bigr)^2, \\
    \mathtt{LLB}_{\phi,\yy_1} & := \min_{\substack{\xx\in\mathcal X_m\\ \sum_{j=1}^{\bar m}(x_j-(m-j+1))^2\le\delta^2}} \sum_{j=1}^{\bar m}(x_j-j)^2.
\end{align*}
Observe that the transformation $\xx' := (m{+}1)\mathbf{1} - \xx$ is its own inverse, applying it twice returns the original vector. Hence, we obtain
\begin{align}
    \label{eq:thm4_1}  \mathtt{UUB}_{\phi,\yy_1} & = \mathtt{UUB}_{\phi,\yy_2}, \\
    \label{eq:thm4_2} \mathtt{LLB}_{\phi,\yy_1} & = \mathtt{LLB}_{\phi,\yy_2}.
\end{align}

\paragraph{Step 3: Pruning condition.}
Because $\mathtt{UB}_{\phi,\yy} \leq \mathtt{UUB}_{\phi,\yy}$ and
$\mathtt{LB}_{\phi,\yy} \geq \mathtt{LLB}_{\phi,\yy}$ by definition, the inequality $\mathtt{UUB}_{\phi,\yy_1} \leq \mathtt{LLB}_{\phi,\yy_2}$ implies $\mathtt{UB}_{\phi,\yy_1} \leq \mathtt{LB}_{\phi,\yy_2}$, so node $\phi$ is pruned
(no ranking vector in $\mathcal{S}_\phi$ can be closer to $\yy_2$ than to $\yy_1$). By symmetry, that is applying \eqref{eq:thm4_1} and \eqref{eq:thm4_2}, the same holds with the roles of the two centroids swapped.

\paragraph{Step 4: Depth bound.} Consequently, the tree can branch beyond level $\bar{m}$ only if $\mathtt{UUB}_{\phi,\yy_1} > \mathtt{LLB}_{\phi,\yy_2}$. Taking the worst-case node on level $\bar{m}$ leads exactly to the optimization problems inside the definition of $\mu(m, \delta)$ in the theorem statement. Hence the tree becomes empty and the reconstruction step terminates, once the level index reaches
\begin{align*}
    \bar{m} & = \mu(m,\delta) := \min\Bigl\{ \bar{m} \mid \mathtt{UUB}_{\phi,\yy_1} \leq \mathtt{LLB}_{\phi,\yy_2} \Bigr \},
\end{align*}
so the depth is at most $\mu(m,\delta)$. This completes the proof. 
\qed

\subsection{Proof of Proposition \ref{pro:bound_on_omega_iterations}}
\textit{Proof. }Let the random variable $\UU_{\ell}$, for $\ell = 0, 1, \ldots$, denote the $\ell^\text{th}$ randomly generated ranking vector in the given iterative procedure, and define $\UU_0 := \yy$. We prove the result by induction, and all subsequent relationships hold almost surely. First, we derive Equation \eqref{eq:u_ell_u_0} and Inequality \eqref{ineq:u_ell_u_o}, which are used in the proof.
For $\ell \geq 1$, we have 
\begin{align}
    \label{eq:u_ell_u_0} \| \UU_{\ell} - \UU_{\ell-1} \|^2 & = 2,
\end{align}
as only two consecutive integers in $\UU_{\ell-1}$ are swapped to form $\UU_{\ell}$. For $\ell \geq 2$, we apply the triangle inequality and proceed to derive::
\begin{align}
    \nonumber \| \UU_{\ell} - \UU_{0} \|^2 & = \| \UU_{\ell} - \UU_{\ell-1} + \UU_{\ell-1} - \UU_{0} \|^2 \\
    \label{ineq:u_ell_u_o} & \leq \left( \| \UU_{\ell} - \UU_{\ell-1} \| + \| \UU_{\ell-1} - \UU_0 \| \right)^2.
\end{align}
For the base case $\ell = 1$, Equation \eqref{eq:u_ell_u_0} shows that the target inequality $\| \UU_{1} - \UU_{0} \|^2 \leq 2(1)^2$ holds. Additionally, for the case $\ell = 2$, \eqref{ineq:u_ell_u_o} and \eqref{eq:u_ell_u_0} together show that 
\begin{align*}
    \| \UU_{2} - \UU_{0} \|^2 & \leq \left( \| \UU_{2} - \UU_{1} \| + \| \UU_{1} - \UU_0 \| \right)^2 \\
    & = \left( \sqrt{2} + \sqrt{2} \right)^2 \\
    & = 2(2)^2,
\end{align*}
confirming that the target inequality holds. Now, let us assume that the target inequality holds for $\ell < \omega$ and prove that it also holds for $\ell = \omega$. From \eqref{ineq:u_ell_u_o}, we have
\begin{align*}
    \| \UU_{\omega} - \UU_{0} \|^2 & \leq \left( \| \UU_{\omega} - \UU_{\omega-1} \| + \| \UU_{\omega-1} - \UU_0 \| \right)^2,
\end{align*}
and using the induction hypothesis, we further obtain
\begin{align*}
    \| \UU_{\omega} - \UU_{0} \|^2 & \leq \left( \sqrt{2} + \sqrt{2}(\omega - 1) \right)^2 \\
    & = \left( \sqrt{2} \omega \right)^2 \\
    & = 2 \omega^2.
\end{align*}
This completes the proof. 
\qed

\end{APPENDICES}


\end{document}